%% file: paper.tex
\PassOptionsToPackage{table}{xcolor}
\documentclass{opendatalab}

\usepackage[utf8]{inputenc}
\usepackage{booktabs}
\usepackage{colortbl} 
\usepackage{hyperref}
\usepackage{graphicx}
\usepackage{multirow}  
\usepackage{graphicx}   
\usepackage{array}         
\usepackage{paracol}    
\usepackage{float}
\usepackage{longtable}
\usepackage{listings}
\usepackage[numbers]{natbib}

\definecolor{codegreen}{rgb}{0,0.6,0}
\definecolor{codegray}{rgb}{0.5,0.5,0.5}
\definecolor{codepurple}{rgb}{0.58,0,0.82}
\definecolor{backcolour}{rgb}{0.95,0.95,0.92}
\definecolor{promptcolor}{HTML}{D1D0F2}
\definecolor{promptcolorheader}{HTML}{bdbcec}

\newcommand{\promptbox}[2]{
\begin{tcolorbox}[
top=0.3em,bottom=0.3em,left=0.5em,right=0.5em,
toptitle=0.3em,bottomtitle=0.2em,boxsep=0pt,
colframe=promptcolorheader,colback=promptcolor!50,boxrule=0.5pt,
]
\footnotesize
\end{tcolorbox}
}
\lstdefinestyle{mystyle}{
    backgroundcolor=\color{backcolour},   
    commentstyle=\color{codegreen},
    keywordstyle=\color{magenta},
    numberstyle=\tiny\color{codegray},
    stringstyle=\color{codepurple},
    basicstyle=\ttfamily\footnotesize,
    breakatwhitespace=false,         
    breaklines=true,                 
    captionpos=b,                    
    keepspaces=true,                 
    numbers=left,                    
    numbersep=5pt,                  
    showspaces=false,                
    showstringspaces=false,
    showtabs=false,                  
    tabsize=2
}

\title{%
  \textcolor{odlblue!40}{E}%
  \textcolor{odlblue!45}{n}%
  \textcolor{odlblue!50}{v}%
  \textcolor{odlblue!55}{i}%
  \textcolor{odlblue!60}{s}%
  \textcolor{odlblue!65}{i}%
  \textcolor{odlblue!70}{o}%
  \textcolor{odlblue!75}{n}%
  : Benchmarking Unified Understanding \& Generation for Causal World Process Insights
}

 \author[1]{Juanxi Tian}
 \author[1]{Siyuan Li}
 \author[1]{Conghui He}
 \author[1]{Lijun Wu}
 \author[1]{Cheng Tan}

 \affiliation[1]{Shanghai Artificial Intelligence Laboratory}
\abstract{
Current multimodal models aim to transcend the limitations of single-modality representations by unifying understanding and generation, often using text-to-image (T2I) tasks to calibrate semantic consistency. However, their reliance on static, single-image generation in training and evaluation leads to overfitting to static pattern matching and semantic fusion, while fundamentally hindering their ability to model dynamic processes that unfold over time. To address these constraints, we propose Envision—a causal event progression benchmark for chained text-to-multi-image generation. Grounded in world knowledge and structured by spatiotemporal causality, it reorganizes existing evaluation dimensions and includes 1,000 four-stage prompts spanning six scientific and humanities domains. To transition evaluation from single images to sequential frames and assess whether models truly internalize world knowledge while adhering to causal-temporal constraints, we introduce Envision-Score—a holistic metric integrating multi-dimensional consistency, physicality, and aesthetics. Comprehensive evaluation of 15 models (10 specialized T2I models, 5 unified models) uncovers: specialized T2I models demonstrate proficiency in aesthetic rendering yet lack intrinsic world knowledge. Unified multimodal models bridge this gap, consistently outperforming specialized counterparts in causal narrative coherence. However, even these unified architectures remain subordinate to closed-source models and struggle to overcome the core challenge of spatiotemporal consistency. This demonstrates that a focus on causally-isolated single images impedes multi-frame reasoning and generation, promoting static pattern matching over dynamic world modeling—ultimately limiting world knowledge internalization, generation.
}

\date{\today}
\correspondence{Cheng Tan, \email{tancheng@pjlab.org.cn}}

\metadata{%
  \makebox[\textwidth][c]{%
    \href{https://github.com/opendatalab-raiser/Envision}{%
      \includegraphics[height=1.2em]{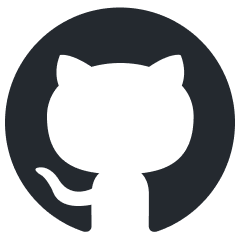}%
      \hspace{0.2em}%  
      \textbf{Code}%
    }%
    \hspace{2em}%
    \href{https://huggingface.co/datasets/opendatalab-raiser/Envision}{%
      \raisebox{-0.7ex}{\includegraphics[height=1.8em]{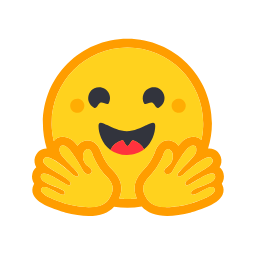}}%
      \textbf{Dataset}%
    }
    \hspace{2em}%
    \href{https://opendatalab-raiser.github.io/Envision/}{\includegraphics[height=1.1em]{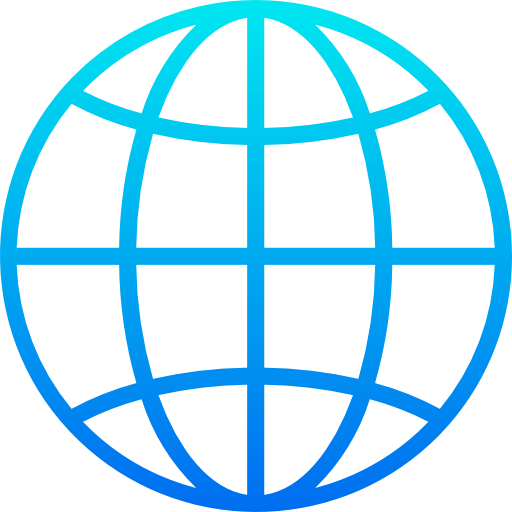} \textbf{Website}}%
  }%
}

\begin{document}

\maketitle

%\tableofcontents

\section{Introduction}
\label{section:intro}
Current text-to-image (T2I) models are capable of rendering images with remarkable realism and diversity \cite{ramesh2021zerot2i, saharia2022photorealistic, podell2023sdxl}. Efforts are also being made to further push the limits of semantic representation by unifying visual understanding and generation capabilities \cite{wu2024vila, dong2023dreamllm, xie2024showo}. Additionally, tasks of T2I benchmarks are used to calibrate the semantic consistency of world knowledge and the physical plausibility of visual representations across models. However, this reliance on static pattern matching of single-frame images may lead to overfitting limitations in the models, where semantics become fused. Inherently, static images lack temporal directionality, creating causal ambiguity where the genesis and consequence of a visual state are indistinguishable. Models trained solely on such single-image data may excel at constructing static scenes, yet they lack the capacity to govern the dynamic processes of real-world events.\cite{chen2025unireal,yang2023learning}. Therefore, the ability to generate photorealistic frames does not necessarily imply an understanding of the generative rules of reality. This paradox compels us to ask: \textbf{Does outstanding performance in generating static, isolated images fully demonstrate that the model has truly internalized world knowledge?}

\begin{figure*}[t]
    \centering
    % 调整图片部分宽度
    \begin{minipage}{0.42\textwidth}
        \centering
        \includegraphics[width=0.95\linewidth]{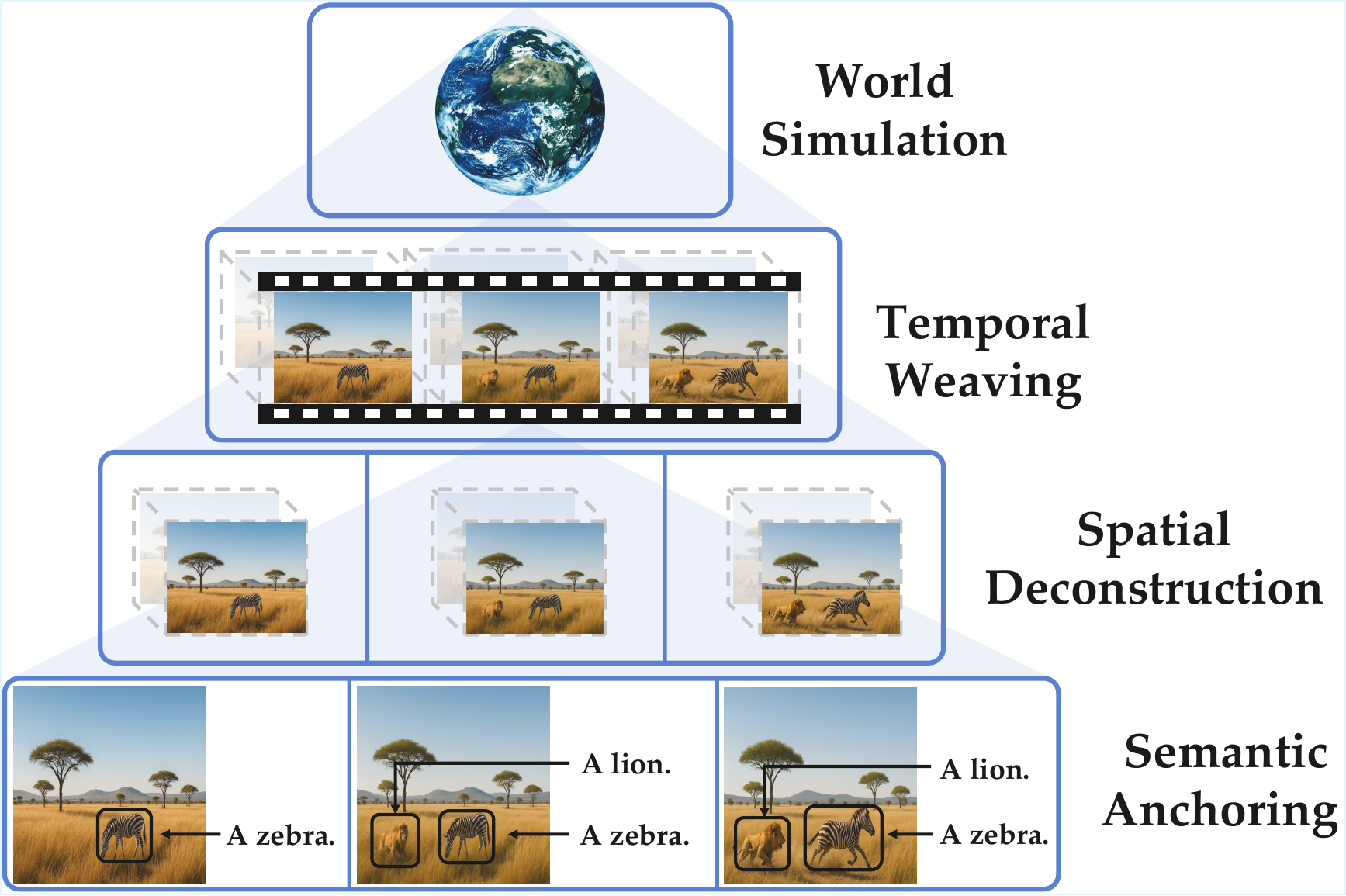}
        % 缩小图片标题字体
        \vspace{-0.50em}\caption{\footnotesize Envision Vision: (1) Semantic Anchoring, (2) Spatial Deconstruction, (3) Temporal Weaving, and (4) World Simulation. Progressive stages of cognitive development in generative models.}
        \label{fig:case}
    \end{minipage}
    \hspace{0.5em}
    % 调整表格部分宽度
    \begin{minipage}{0.55\textwidth}
        \centering
        \captionof{table}{Requirements for Different Text-to-Visual Generation Modalities.}
        \label{tab:requirements}
        \vspace{-0.75em}
        \scriptsize
        \begin{tabular}{l>{\raggedright\arraybackslash}p{2.8cm}>{\raggedright\arraybackslash}p{3.2cm}}
        \toprule
        \textbf{Modality} & \textbf{Core Requirements} & \textbf{Additional Requirements} \\
        \midrule
        \textbf{T2I} & 
        \hspace{0.1em}Image Aesthetics\newline
        \hspace{0.5em}Object Texts\newline
        \hspace{0.5em}Position Texts & ---
        \\
        \midrule
        % T2V 行已删除
        \textbf{T2I to T2MI} & 
        All T2I Requirements &
        + Chain of Events\newline
        + Consistent Attributes\newline
        + Event Causality \\
        \midrule
        \textbf{T2MI to T2V} & 
        All T2MI Requirements &
        + Chain of Actions\newline
        + Object \& Attribute Consistency\newline
        + Temporal Continuity \\
        \bottomrule
        \end{tabular}
        
        \vspace{0.1em}
        \raggedright
        \scriptsize
        \textbf{Note:} T2I: Text-to-Image; T2V: Text-to-Video; T2MI: Text-to-Multi-Image. Core requirements are fundamental capabilities; `+` denotes extensions beyond preceding modalities.
    \end{minipage}
    \vspace{-1.50em}
\end{figure*}

Building on this, our idealized vision posits that a perfected generative model must possess the capacity to truly internalize and govern world knowledge, enabling it to construct an internal world model through the dynamic interleaving of understanding and generation (as illustrated in Figure \ref{fig:case}). This vision, termed \textbf{Envision-Vision}, is concretely embodied in four stages: (1) Semantic Anchoring (mapping visual features to conceptual entities), (2) Spatial Deconstruction (deconstructing 2D spatial relationships using implicit 3D cognition), (3) Temporal Weaving (constructing cross-temporal causal chains and state transitions), and (4) World Simulation (building an internal world model to predict spatiotemporal states and simulate process dynamics).

Under the \textbf{Envision-Vision}, various visual modalities are intricately interconnected, ultimately advancing toward a comprehensive visual intelligence capable of representing the complete world. As an optimal intermediate representation, multi-image sequences bridge the spatiotemporal boundary between images and videos, employing continuous or discrete image sequences to depict event processes under arbitrary spatiotemporal constraints, as detailed in Table \ref{tab:requirements}. In summary, we introduce Envision—a causal event progression benchmark built upon world knowledge, centered on causal spatiotemporal constraints, and generating multi-image sequences from chained textual descriptions. We meticulously design \textbf{Envision-Score}, a quantitative evaluation metric integrating multidimensional aesthetic, consistency, and physical considerations, compelling models to simulate coherent causal trajectories rather than merely matching static patterns. This prompts profound reflection on the transition from individual event frames to the entire process of the event.
%with a particular focus on models’ performance in generating both per-frame and overall sequence coherence for events.

Envision evaluates 5 UMMs and 10  specialize T2I models across 1,000 four-stage prompt sequences spanning six subdomains in natural sciences and humanities history. The benchmark assesses whether models can genuinely internalize the world knowledge introduced during training and maintain the capabilities reflected in previous benchmarks, within dynamically constrained causal spatiotemporal ``event processes.'' In Envision, understanding and generation are tightly coupled: failures in generating any event frame within the overall event sequence not only highlight the limitations of prior static single-image benchmarks but also reveal disconnections or conflicts between these two capacities in UMMs. These critical shortcomings expose the substantial gap between ambitious ideals and actual capabilities. Our main contributions are as follows:
\begin{itemize}
    \item \textbf{\textit{Novel Benchmark.}} We introduce \textbf{Envision}, a novel benchmark designed to evaluate T2I models' ability to produce dynamic causal event sequences through multiple image sequences.
    
    \item \textbf{\textit{Event-Level Evaluation.}}  We propose \textbf{Envision-Score}, a dedicated metric for comprehensively evaluating event-level multi-image sequences across aesthetics, consistency, and physical.
    
    \item \textbf{\textit{Analysis and Insights.}} We reveal the performance gap between current T2I models in static, isolated single-image generation and dynamic, causal reasoning-based multi-image generation.
\end{itemize}

\vspace{-0.75em}
\section{Why the Multiple-Image Sequence?}
\begin{figure*}[!htbp]
\vspace{-0.50em}
\centering
\includegraphics[width=\textwidth]{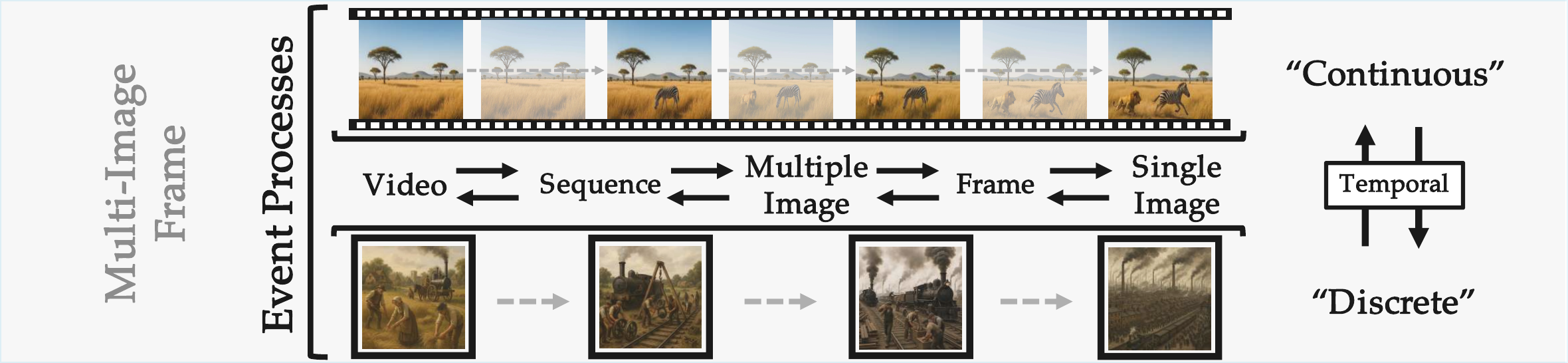}
\caption{Overview of Envision's Multi-Image Framework.}
\label{fig:mi}
\vspace{-1.50em}
\end{figure*}
\subsection{Multi-Image as a Flexible Framework for Visual Narrative Construction.} 
The multi-image framework (Figure \ref{fig:mi}) introduced by Envision establishes a flexible framework for modeling event processes across diverse spatiotemporal constraints. By moving beyond single-image generation—which captures only isolated moments—this approach enables the construction of coherent visual narratives that seamlessly adapt to both continuous and discrete event representations. In continuous temporal evolution, models are required to generate smooth, physically consistent transitions that adhere to conservation laws, while in discrete contexts, they must preserve logical progression despite leaps in temporal resolution. This dual capacity facilitates rigorous evaluation of a model’s ability to handle both fine-grained dynamic processes and high-level event sequencing. Spatially, the methodology offers complementary versatility: sequences can maintain consistent viewpoints to document progressive changes within fixed scenes, or employ varying perspectives to trace developments across different spatial contexts. This spatial adaptability, combined with temporal flexibility, enables the representation of intricate world processes characterized by complex spatiotemporal interactions. For a detailed description, please refer to the Appendix \ref{app:event-structure}.
%As a result, the Envision benchmark provides a more holistic and nuanced assessment of scene modeling capabilities than conventional single-image evaluations permit, effectively bridging the gap between static visual understanding and dynamic causal reasoning.

\subsection{Bidirectional Verification Between Understanding and Generation.} The Envision establishes a bidirectional evaluation paradigm for T2I models, rigorously examining the interaction between understanding and generation within UMMs. It constructs visual narrative scenes through progressively generated image outputs. This serves to demonstrate whether the UMM outperforms specialized T2I models in internalizing world knowledge and generating expression, and whether understanding and generation are truly unified, dynamically interleaved, and mutually reinforcing. In the forward direction of \textbf{($\text{Understanding} \rightarrow \text{Generation}$)}, a model's internalized world knowledge and causal reasoning capabilities are systematically probed. This process mandates the generation of a causally coherent multi-image sequence from an abstract conceptual understanding, and must strictly adhere to spatiotemporal and physical constraints. The occurrence of an interruption in the sequence generation—such as producing an illogical state transition or violating physical laws—constitutes a critical diagnostic indicator, revealing a fundamental deficiency in the model's capacity to internalize world knowledge and deploy it for visual generative expression.

Conversely, \textbf{($\text{Generation} \rightarrow \text{Understanding}$)} highlights how the act of generation itself functions as an analytical mechanism that deepens understanding. Whole process is fundamentally an interleaving cycle: When confronted with ambiguities or inconsistencies during sequential frame generation, the model must internally refine its causal representations based on this visual state. By treating visual generation as the result of interleaved reasoning processes of understanding and generation, this approach examines the model's dynamic conceptual and structural comprehension of an event sequence during its step-by-step generation. This enables analysis of whether the integration of understanding and generation within UMMs exhibits disconnects or conflicts, and whether this architectural paradigm offers advantages over specialized T2I models.
\begin{figure*}[t!]
\centering
\includegraphics[width=0.94\textwidth]{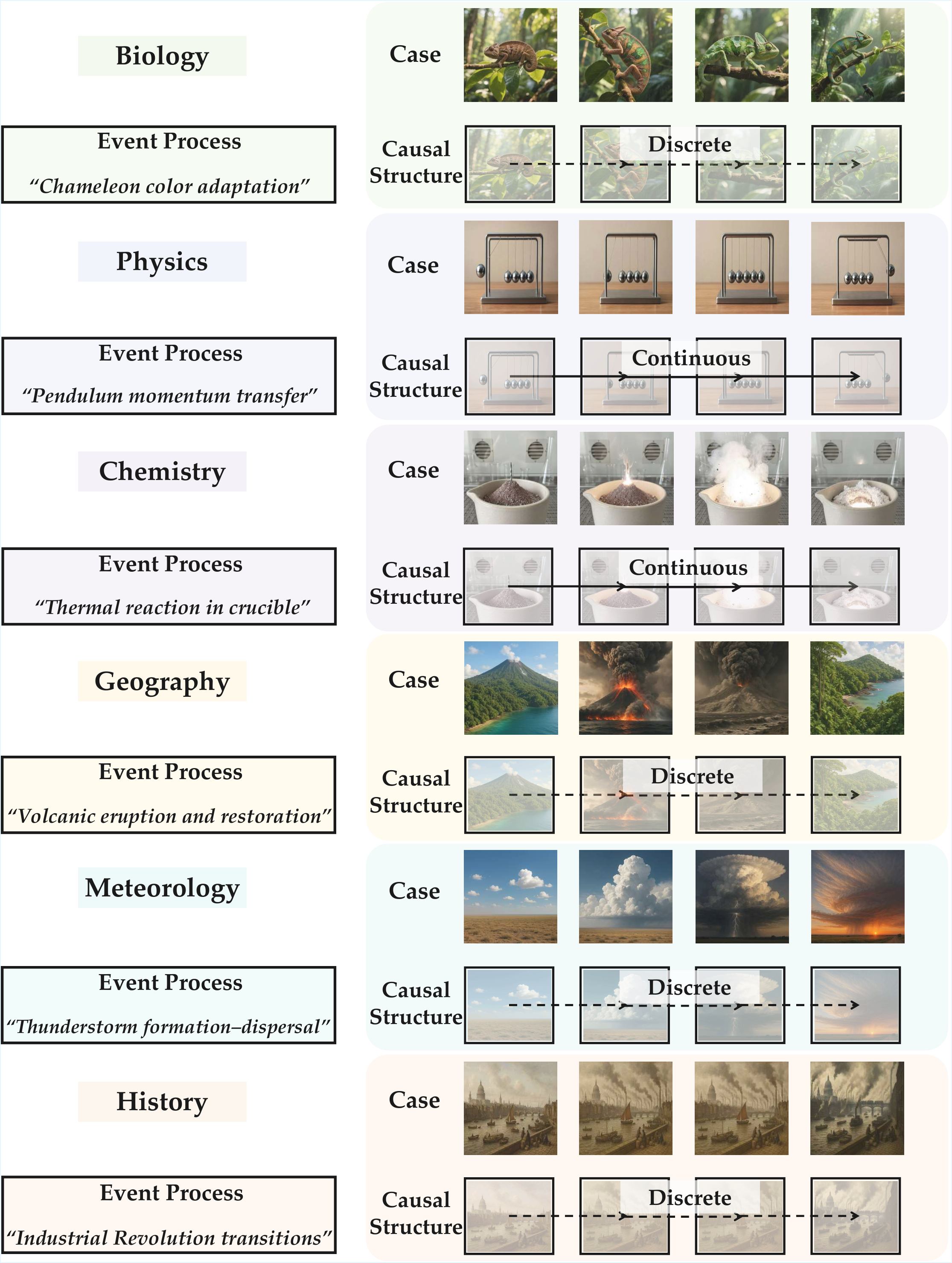} % 缩小为原宽度的85%
\caption{
    Spanning six disciplines, representative cases showcase the causal structure of processes as discrete (dashed) or continuous (solid) spatial relations over time.
}
\label{fig:case_full}
\end{figure*}
\section{Envision}
\label{sec:method}
\begin{figure*}[t]
    \centering
    % \vspace{-1.50em}
    \includegraphics[width=\textwidth]{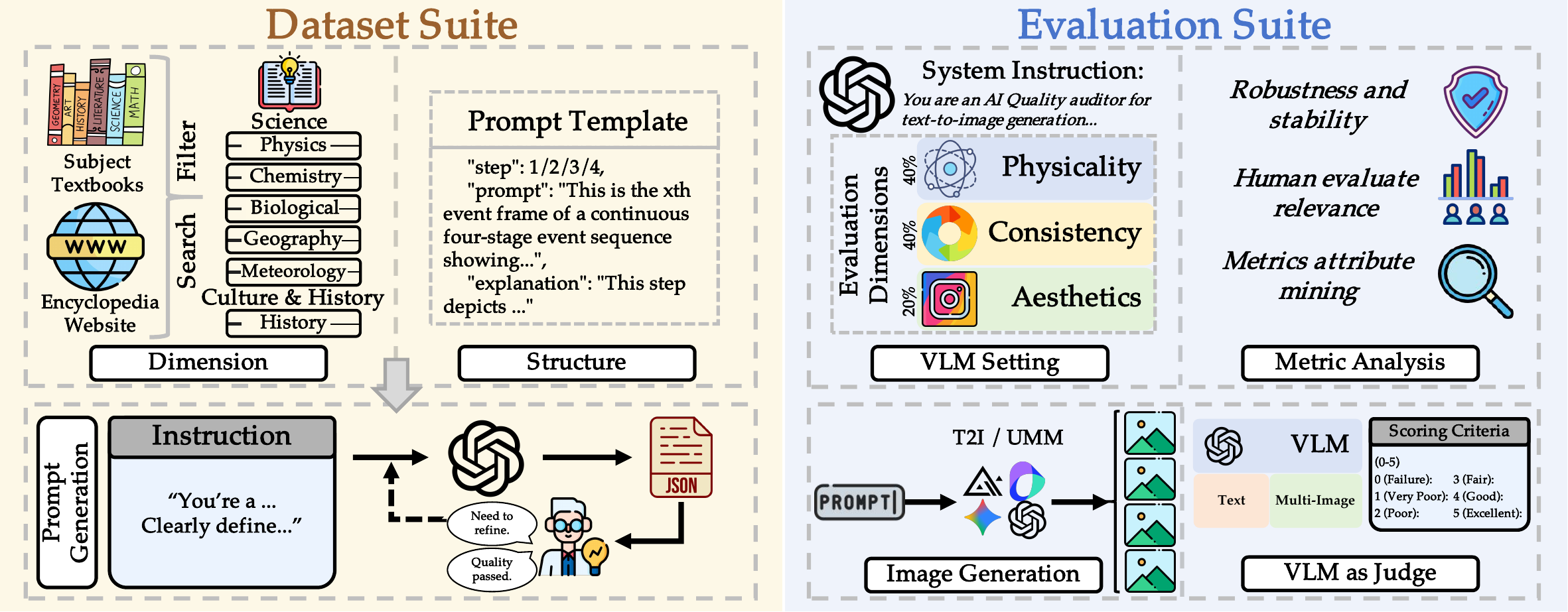}
    \caption{Overview of the Event Process in generating and evaluating multi-image sequences for Envision. The process involves constructing a Dataset Suite using multi-domain knowledge sources (Science, Culture \& History) and structured prompt generation. This feeds into an Image Generation process via a Text-to-Image model, followed by an Evaluation Suite which includes both Physicality, consistency, and aesthetics, and VLM as Judge and Metric Analysis using specific scoring criteria (0-5).}
    \label{fig:pipeline}
    \vspace{-1.00em}
\end{figure*}

Envision is founded on a fundamental premise: genuine visual intelligence must transcend the limitations of static isolation to encompass the understanding and simulation of dynamic world processes. Current T2I evaluation paradigms, predominantly focused on single-image tasks, fail to capture the causal relationships and spatiotemporal continuity inherent in real-world phenomena. To address this limitation, we propose Envision—a comprehensive benchmark centered on multi-image generation that compels models to generate the event process image by image, thereby demonstrating their ability to represent the true event sequences of the world.

\subsection{World Knowledge Dataset Suite}
\begin{figure*}[t!]
\centering
\vspace{-1.00em}
\includegraphics[width=\textwidth]{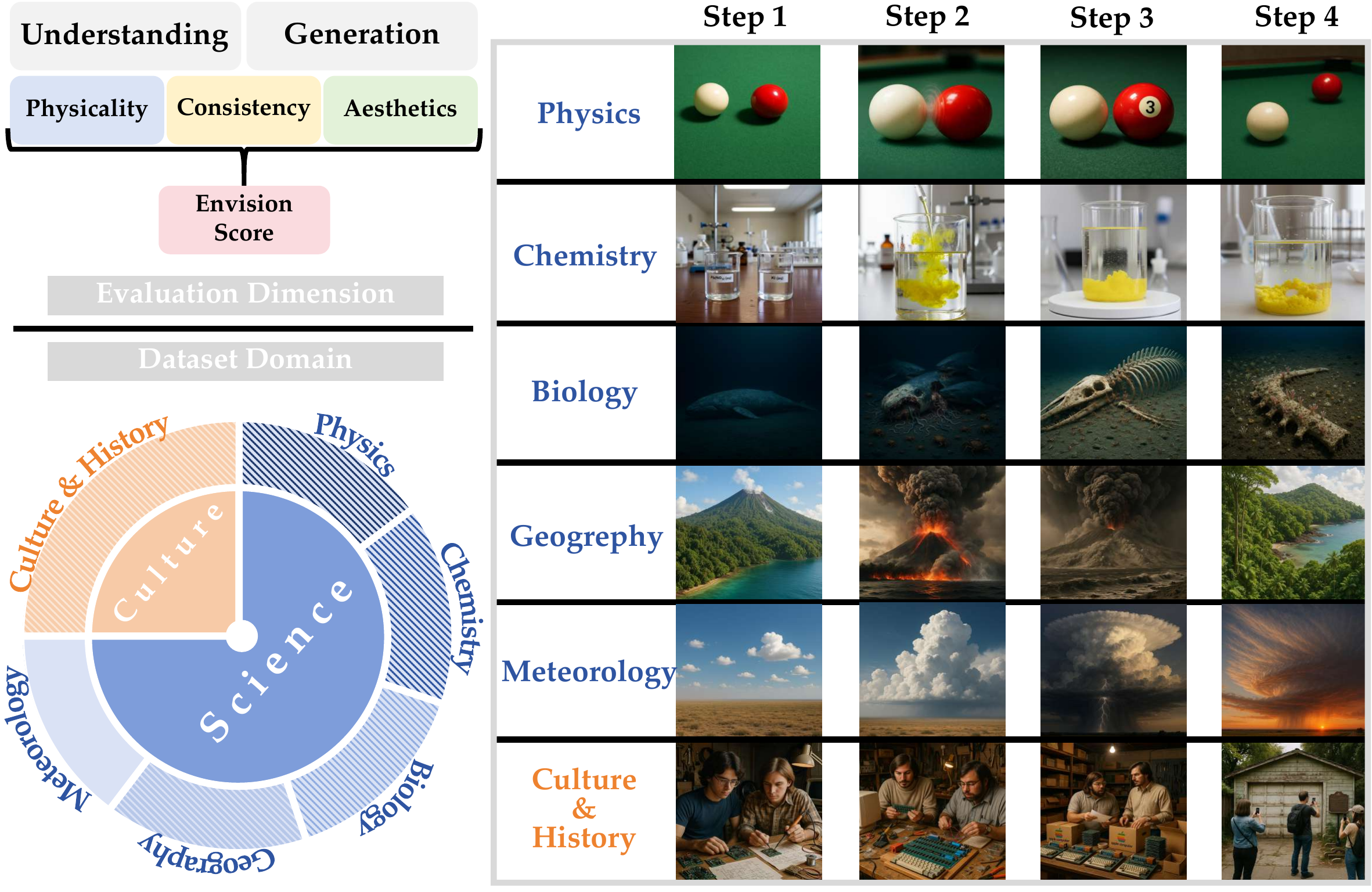}
\caption{Overview of the Envision Data Domain \& Evaluation Dimension.}
\label{fig:domain}
\vspace{-1.50em}
\end{figure*}
Envision is constructed on two core domains—natural sciences and humanities history—drawing on carefully curated information from academic textbooks and online sources that center on real-world events, covering a total of six disciplines. To balance the difficulty of generating visual narratives with the pressure of context length in evaluation, we employ a sequential prompt template consisting of four-stage prompts. Under human experts supervision, pre-collected and filtered information was utilized by GPT-4o to generate and refine four-stage narrative prompts for each event. This process ultimately produced 1,000 high-quality event sequences, comprising 4,000 text-to-image prompts, as illustrated in Figure \ref{fig:domain} and Figure \ref{fig:case_full}.

\subsubsection{Nature Science}
This category evaluates the model's internalization and representation of natural science phenomena and fundamental laws, comprising 75\% of the dataset. Its five sub-dimensions each contain 150 sequences. For specific examples, see Figure \ref{fig:sci}.
\begin{itemize}
    \item \textbf{Physics:} The tasks in this domain aim to examine the model's reasoning capabilities regarding core principles such as mechanics, thermodynamics, and electromagnetism.

    \item \textbf{Chemistry:} Assess the level of understanding of fundamental chemical principles, including chemical reaction kinetics, molecular interactions, and phase transitions, etc.

    \item \textbf{Biology:} Evaluations focus on biological processes such as ecological succession, predator-prey dynamics, organismal life cycles, and evolution.

    \item \textbf{Meteorology:} Focusing on dynamic atmospheric systems, specific process deductions are made based on principles of meteorology and thermodynamics to infer the development and changes in space and time.
    
    \item \textbf{Geography:} Study geology, geomorphology and environmental systems. Its tasks include geological processes such as river meandering and coastal erosion, and reasoning about the topographic evolution driven by factors such as erosive forces, tectonic activities.
\end{itemize}

\subsubsection{History \& Cultural}
\begin{figure*}[t!]
\centering
\begin{subfigure}[b]{1.0\textwidth}
    \includegraphics[width=\textwidth]{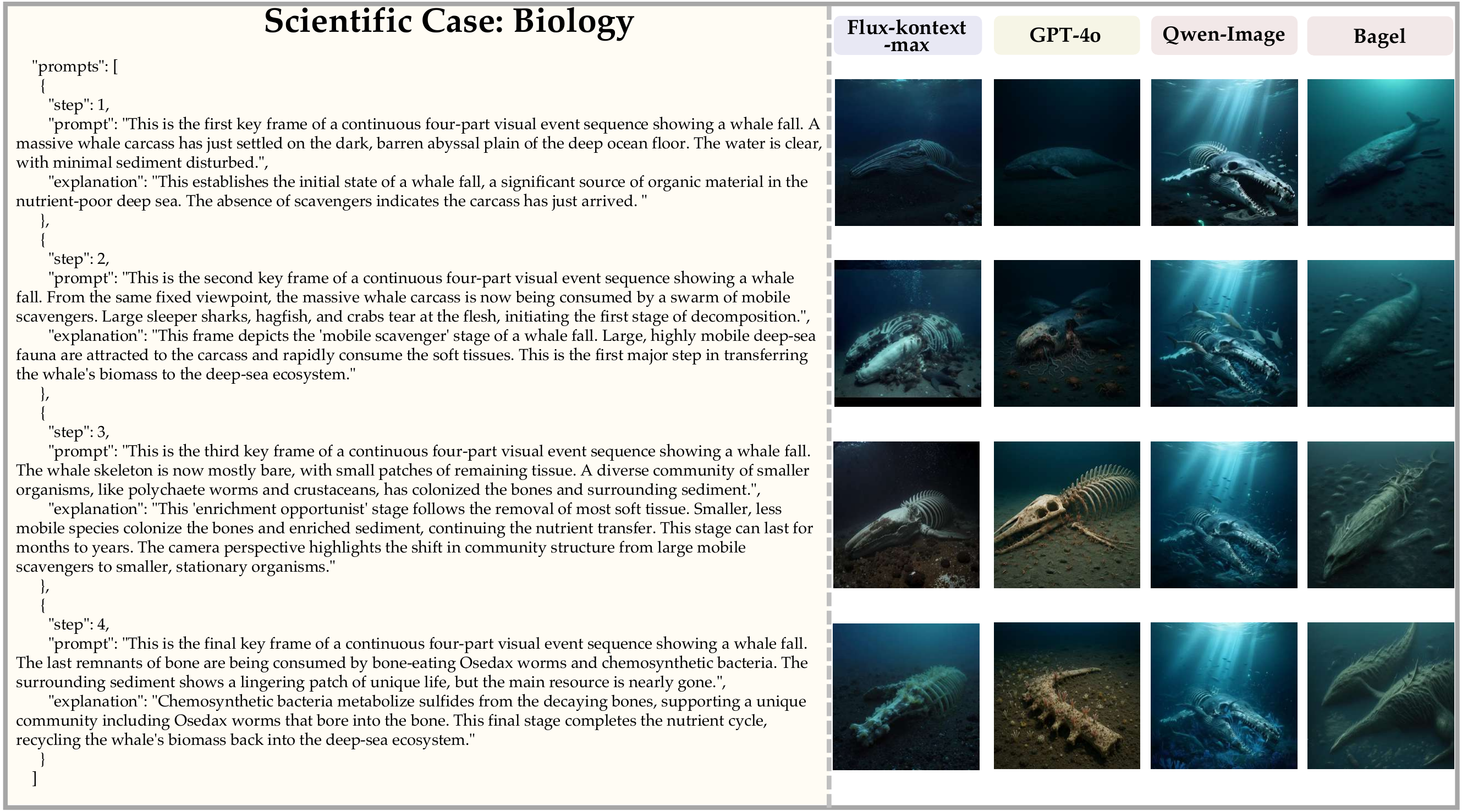}
    \subcaption{A four-stage visual narrative of a whale fall event in the deep ocean.}
    \label{fig:sci}
\end{subfigure}
\\
\begin{subfigure}[b]{1.0\textwidth}
    \includegraphics[width=\textwidth]{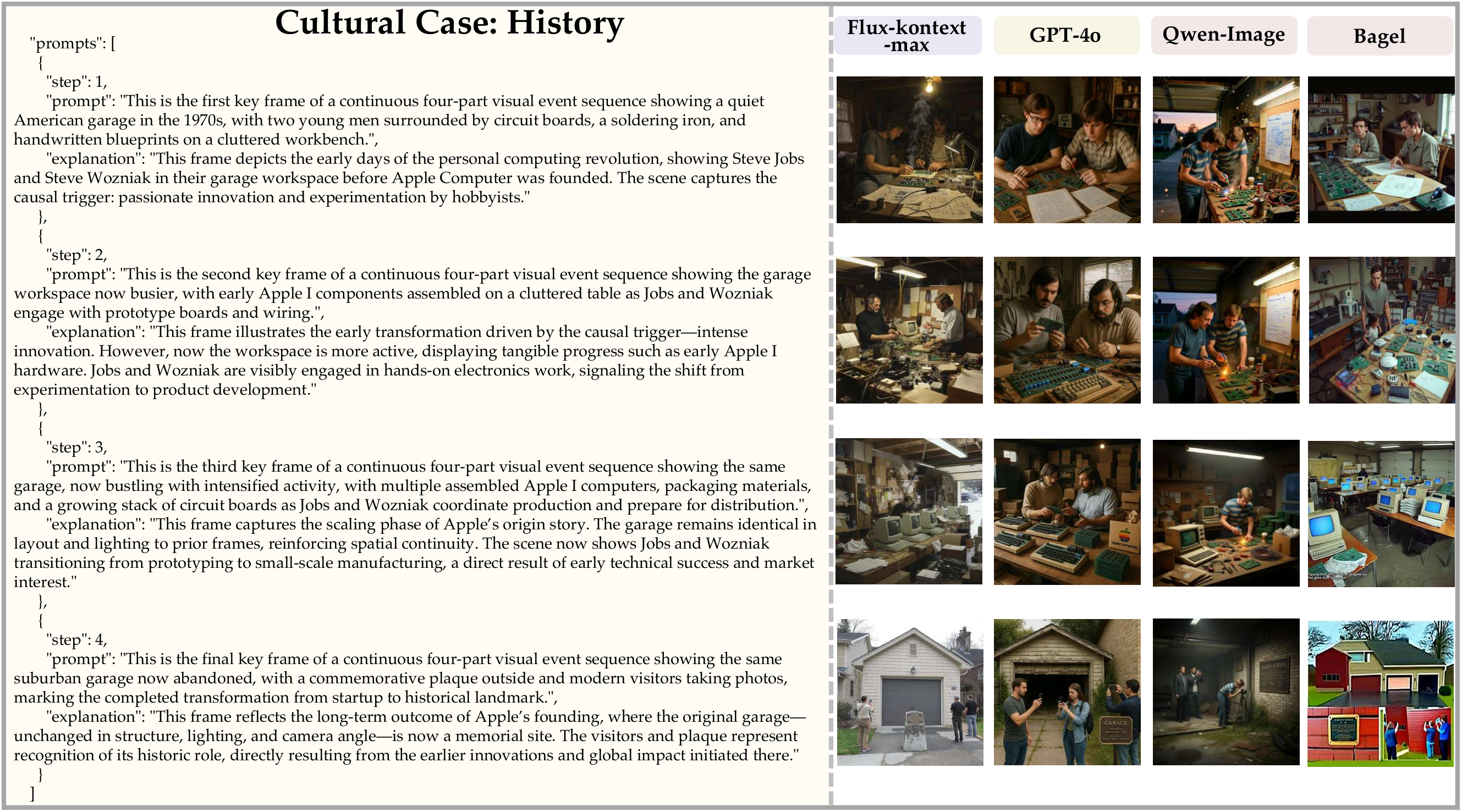}
    \subcaption{A four-stage historical narrative showing the evolution of Apple Computer's founding in a 1970s garage.}
    \label{fig:cul}
\end{subfigure}
\caption{Exemplar four-stage visual narratives from the Envision benchmark.}
\label{fig:combined}
\end{figure*}

This category encompasses human social development and historical evolution. Grounded in the context of human history and cultural knowledge, it collectively accounts for one-quarter of the total data volume. For specific examples, please refer to \ref{fig:cul}.
\begin{itemize}
    \item \textbf{World History \& Cultural Commonsense:} This category requires models not only to comprehend the human intent and social logic underlying event processes, but also to discern the intricate connections between scientific and technological advancements and sociocultural phenomena during periods of social transformation.
\end{itemize}

\subsubsection{Causal Structure}
Across all domains, Envision sequences are fundamentally grounded in temporal causality, requiring models to generate events that unfold logically over time. However, to rigorously test different facets of world modeling, we stratify these sequences into two distinct spatial-temporal structures: \textbf{Continuous} and \textbf{Discrete} (Figure\ref{fig:case_full}).

\begin{itemize}
    \item \textbf{Continuous Causality:} These sequences maintain a consistent spatial context with smooth, uninterrupted temporal progression, typical of fine-grained physical dynamics (\textbf{e.g.,} chemical reactions, immediate mechanical processes). This structure compels models to strictly adhere to physical conservation laws and accurately simulate subtle state transitions.
    \item \textbf{Discrete Causality:} These sequences involve significant spatial leaps or substantial temporal jumps between frames, commonly found in macro-scale processes (\textit{e.g.,} geological evolution, multi-stage life cycles, historical narratives). This structure necessitates abstracting high-level logic and maintaining long-range coherence across discontinuous contexts, evaluating its capacity for abstract causal reasoning.
\end{itemize}

%This dual-structure design ensures that Envision evaluates not only a model's ability to simulate immediate physical dynamics but also its capacity to sustain long-range semantic coherence across spatially disjointed events.
\subsection{Evaluation Suite}
Envision's evaluation is built upon a meticulously designed workflow, aiming to achieve an objective and reasonable assessment of multi-image and text pairs that constitute event-level data. To ensure objectivity, reproducibility, and scoring efficiency, moving beyond subjective human judgments, we adopt a composite methodology centered on three pillars: (1) Event Sequence-level Evaluation, (2) Deconstruction and Integration of Evaluation Dimensions, and (3) VLM-as-Judge Evaluation.

\vspace{-0.75em}
\subsection{Evaluation Dimensions}
To fulfill the stringent requirements of event-level visual narrative, our evaluation framework is founded upon three necessary dimensions: Consistency, Physicality, and Aesthetics. Furthermore, to ensure comprehensive and granular reliability, particularly given the inherent limitations of VLM-as-Judge models in differentiating subtly similar criteria, we meticulously delineate each dimension into distinct, non-overlapping sub-metrics. The specific evaluation criteria are as follows:
\begin{itemize}
    \item \textbf{Consistency:} Ensuring an unbroken multi-image sequence, this metric evaluates the preservation of logical, factual, and narrative coherence throughout the sequence. Sub-dimensions include: \textbf{Semantic Consistency}, \textbf{Spatial-Temporal Consistency}, and \textbf{Factual Consistency}.
    
    \item \textbf{Physicality:} Focused on the plausibility of dynamic processes, this metric quantifies a model's internalization of physical laws and its capacity for reliable simulation. Sub-dimensions include: \textbf{Basic Properties}, \textbf{Dynamics and Interactivity}, \textbf{Physical Reliability}.
    
    \item \textbf{Aesthetics:} Ensure that narrative expression does not come at the expense of aesthetic quality. Sub-dimensions include: \textbf{Expressiveness}, \textbf{Aesthetic Quality}, and \textbf{Authenticity}.
\end{itemize}
\vspace{-0.75em}
\subsection{Scoring Rubric}
Each metric is scored by GPT-4o on a discrete integer scale from 0 to 5. This discrete scale serves to mitigate both extreme scoring tendencies and central tendency bias, thereby ensuring a statistically discriminative distribution of scores. This rubric provides a standardized measure of quality, with clear criteria for each score level to ensure inter-rater reliability.

\begin{itemize}
    \item \textbf{5 (Excellent):} The generated sequence of images is flawless regarding the specific sub-dimension. It perfectly adheres to the prompt and the principles of narrative.
    \item \textbf{4 (Good):} The generated sequence of images exhibits only minor, inconsequential deviations. The core causal logic and visual intent are fully preserved.
    \item \textbf{3 (Fair):} There are noticeable errors, but they do not fundamentally violate the causal chain or the primary objective of the prompt.
    \item \textbf{2 (Poor):} Contains significant errors that undermine the causal narrative or visual consistency. The generated state transition is illogical or physically implausible.
    \item \textbf{1 (Very Poor):} The generated sequence of images are fundamentally flawed, bearing little resemblance to a causally or semantically correct outcome.
    \item \textbf{0 (Failure):} The model fails to produce a relevant output, or the output is completely nonsensical, artifact-ridden, or non-compliant with the task.
\end{itemize}
\vspace{-0.75em}
\subsubsection{Envision-Score}
During the evaluation process, GPT-4o serves as the scoring model, processing input pairs composed of four-stage narrative prompts and model-generated multi-image sequences for t2I evaluation. Based on the predefined evaluation dimensions and scoring criteria, scores are calculated for three evaluation dimensions—consistency, physicality, and aesthetics—along with their respective sub-metrics. These scores are then consolidated through a weighted average calculation to derive the Envision score.

\paragraph{Weighted Scoring Formulation.}
Let $\mathcal{D} = \{C, P, A\}$ denote the set of primary evaluation dimensions, where $C$ represents \textbf{Consistency}, $P$ represents \textbf{Physicality}, and $A$ represents \textbf{Aesthetics}. Each primary dimension $d \in \mathcal{D}$ comprises a set of sub-dimensions $S_d = \{s_{d,1}, s_{d,2}, \dots, s_{d,n_d}\}$, where $n_d$ is the number of sub-dimensions for dimension $d$, and each $s_{d,i} \in [0, 5]$ represents the score for the $i$-th sub-dimension.

The score for each primary dimension is computed as the arithmetic mean of its constituent sub-dimensions:
\vspace{-0.25em}
\begin{equation}
\mathcal{S}_d = \frac{1}{|S_d|} \sum_{i=1}^{|S_d|} s_{d,i} \quad \text{for } d \in \mathcal{D}
\end{equation}
\vspace{-0.25em}
This method ensures equal weighting within each dimension, reflecting our assessment that all sub-dimensions within Consistency and Physicality contribute equally to their respective constructs, and similarly for Aesthetics.

The comprehensive \textbf{Envision (Overall) Score}, $\mathcal{S}_{\text{Overall}}$, is then computed as a weighted combination of the three primary dimension scores:
\vspace{-0.25em}
\begin{equation}
\mathcal{S}_{\text{Overall}} = \beta_C \cdot \mathcal{S}_C + \beta_P \cdot \mathcal{S}_P + \beta_A \cdot \mathcal{S}_A
\end{equation}
\vspace{-0.25em}
where the weighting coefficients are defined as $\beta_C = 0.4$, $\beta_P = 0.4$, and $\beta_A = 0.2$, satisfying the constraint $\beta_C + \beta_P + \beta_A = 1$.

This formulation explicitly prioritizes causal reasoning by assigning 80\% of the total weight to Consistency and Physicality dimensions, while maintaining a balanced assessment of visual quality through the Aesthetics dimension.

\paragraph{Multi-trial Evaluation Protocol.} To ensure the statistical reliability of our automated evaluation framework, we implement a comprehensive multi-trial evaluation protocol. For each generated sequence comprising the quadruple $\{(I_t, P_t\}_{t=1}^4$, where $I_t$ denotes the image at step $t$, $P_t$ represents the corresponding prompt, we conduct $K$ independent evaluation trials under identical conditions.

For each trial $k \in \{1, 2, \dots, K\}$, the evaluation framework produces a complete evaluation tuple:
\vspace{-0.75em}
\begin{equation}
T_k = (\mathbf{s}_k, r_k)
\end{equation}
\vspace{-0.75em}
where:
\begin{itemize}
    \item $\mathbf{s}_k \in [0,5]^{9}$ represents the fine-grained score vector across all nine sub-dimensions.
    \item $r_k$ denotes the complete textual rationale generated by the evaluator model.
\end{itemize}

This multi-trial data collection enables a dual-faceted analysis of evaluator reliability. First, we perform \textbf{metric analysis} by computing the empirical mean and standard deviation (or upper/lower bounds) for each sub-dimension score across the $K$ trials:
\vspace{-0.75em}
\begin{align}
\mu_{s_{d,i}} &= \frac{1}{K} \sum_{k=1}^K s_{d,i}^{(k)}
\end{align}
\vspace{-1.75em}
\paragraph{Metric Analysis for Evaluation.} Furthermore, we engaged five domain human experts (all holding PhDs) and models from the Gemini\cite{comanici2025gemini}, GPT\cite{hurst2024gpt}, and Qwen\cite{bai2025qwen2,yang2025qwen3} families for evaluation.Each expert selected 50 prompt sequences for each data dimension, with GPT-4o generating corresponding images. The resulting image-text pairs underwent five rounds of evaluation by VLMs and human experts. After analyzing metric outcomes and validating output visualizations, GPT-4o was selected as the automated evaluation tool due to its highly consistent alignment with human expert judgments and robust, stable distribution. This minimizes random fluctuations and ensures assessment consistency. For detailed information, please refer to the Appendix \ref{sec:details} and Figure \ref{fig:model_performance_comprehensive}.

\section{Experiments}
\label{sec:experiments}
\subsection{Experiment Setup}
Our experimental evaluation encompasses 15 models, categorized into three groups: 8 Open-Source T2I Models, 2 Closed-Source T2I Models and 5 UMMs.

\begin{itemize}
    \item \textbf{Open-Source T2I Models:} The open-source T2I model group includes Stable-Diffusion-3.5-flash, Stable-Diffusion-3.5-medium, Stable-Diffusion-3.5-large\cite{podell2023sdxl}, FLUX-dev, FLUX-pro-1.1, FLUX-pro-1.1-ultra, FLUX-kontext-pro and FLUX-kontext-max\cite{batifol2025flux}.
    \item \textbf{Closed-Source T2I Models:} The closed source T2I model group includes GPT-4o\cite{hurst2024gpt}, Gemini-2.5-Flash-Image (nano-banana)\cite{comanici2025gemini}.
    \item \textbf{Unified Multimodal Models (UMMs):} The UMMs under evaluation are Janus-Pro-7B\cite{chen2025januspro}, Hunyuan Image 3.0\cite{cao2025hunyuanimage}, Bagel\cite{deng2025emerging}, Seedream 4.0\cite{seedream2025seedream}, Qwen-Image\cite{wu2025qwen}.
\end{itemize}
\vspace{-0.75em}
\paragraph{Evaluation Protocol.}
For the ``VLM-as-Judge'' component of our evaluation pipeline (see Section~\ref{sec:method}), we employed \textbf{GPT-4o} as the scorer. To ensure fair and reproducible comparisons, image generation was performed using the official default configuration for each model with a fixed random seed. All experiments were conducted on a cluster of eight NVIDIA A800 GPUs.
\input{tables/benchmark}
\begin{table*}[htbp]
\centering
\caption{Comparison of performance in each dimension of the T2I models (All Domains Average).}
\label{tab:comparison_average}
\vspace{-0.75em}
\small
\setlength{\tabcolsep}{1.1mm} 
\begin{tabular}{lcccccccccccc}
\toprule
\multirow{2}{*}{\textbf{Model}} & \multicolumn{12}{c}{\textbf{Evaluation Dimensions}} \\
\cmidrule(lr){2-13}
 & \textbf{SC} & \textbf{FC} & \textbf{STC} & \textbf{CS} & \textbf{Exp} & \textbf{AQ} & \textbf{Auth} & \textbf{AS} & \textbf{PR} & \textbf{BP} & \textbf{DI} & \textbf{PS} \\
\midrule
\rowcolor{blue!5}
\textbf{Open-Source T2I Models} & & & & & & & & & & & & \\
FLUX-dev & 54.11 & 56.30 & 40.98 & 50.37 & 71.05 & 75.66 & 53.02 & 66.60 & 45.23 & 58.97 & 45.84 & 49.96 \\
FLUX-pro-1.1 & 55.20 & 55.89 & 40.56 & 50.45 & 72.95 & 76.73 & 52.67 & 67.30 & 46.96 & 58.18 & 47.62 & 50.88 \\
FLUX-pro-1.1-ultra & 52.04 & 54.53 & 40.71 & 49.01 & 68.69 & 72.68 & 50.29 & 63.75 & 44.88 & 58.28 & 45.82 & 49.61 \\
FLUX-kontext-pro & 57.09 & 60.33 & 48.46 & 55.22 & 64.63 & 70.09 & 56.10 & 63.53 & 52.70 & 66.04 & 49.94 & 56.19 \\
FLUX-kontext-max & 60.85 & 63.46 & 46.06 & 56.48 & 68.42 & 71.94 & 55.83 & 65.30 & 51.24 & 64.38 & 49.07 & 54.86 \\
SD-3.5-flash & 48.12 & 46.05 & 36.56 & 42.44 & 62.92 & 71.76 & 48.08 & 60.79 & 41.08 & 52.20 & 39.87 & 44.35 \\
SD-3.5-large & 40.82 & 50.52 & 33.84 & 43.30 & 63.02 & 67.32 & 45.89 & 58.62 & 41.01 & 53.64 & 40.73 & 45.08 \\
SD-3.5-medium & 49.79 & 50.89 & 35.64 & 44.08 & 61.65 & 67.86 & 48.86 & 59.35 & 41.66 & 54.88 & 42.29 & 46.23 \\
\rowcolor{yellow!8}
\textbf{Closed-Source T2I Models} & & & & & & & & & & & & \\
GPT-4o & \textbf{75.76} & \textbf{78.65} & \textbf{67.42} & \textbf{73.88} & 77.05 & 81.37 & \textbf{71.81} & \textbf{76.70} & \textbf{70.56} & \textbf{79.90} & \textbf{66.44} & \textbf{72.28} \\
Gemini-2.5-Flash-Image & 69.12 & 73.20 & 58.71 & 66.92 & 73.29 & 75.65 & 64.10 & 70.95 & 62.68 & 74.45 & 59.50 & 65.52 \\
\rowcolor{red!5}
\textbf{Unified Multimodal Models} & & & & & & & & & & & & \\
Seedream 4.0 & 66.15 & 66.78 & 56.79 & 63.18 & 73.54 & 76.12 & 58.61 & 69.32 & 60.05 & 69.83 & 56.91 & 62.24 \\
Qwen-Image & 61.16 & 59.88 & 53.20 & 58.03 & \textbf{79.54} & \textbf{82.57} & 58.05 & 73.23 & 55.97 & 67.11 & 54.69 & 59.22 \\
Hunyuan Image 3.0 & 57.42 & 58.14 & 46.03 & 53.78 & 73.08 & 76.05 & 53.48 & 67.40 & 49.54 & 61.34 & 50.70 & 53.81 \\
Bagel & 58.61 & 57.50 & 51.11 & 55.70 & 65.84 & 69.17 & 48.89 & 61.17 & 51.94 & 61.94 & 46.11 & 53.32 \\
Janus-Pro-7B & 52.29 & 55.05 & 41.51 & 49.54 & 63.26 & 64.86 & 45.96 & 57.90 & 44.40 & 53.64 & 44.63 & 47.52 \\
\bottomrule
\end{tabular}

\vspace{0.25em}
\raggedright % 左对齐note
\footnotesize
\textbf{Note:} Evaluation dimension abbreviations: SC: Semantic Consistency, FC: Factual Consistency, STC: Spatial-Temporal Consistency, CS: Consistency Score, Exp: Expressiveness, AQ: Artistic Quality, Auth: Authenticity, AS: Aesthetic Score, PR: Physical Reliability, BP: Basic Properties, DI: Dynamics \& Interactivity, PS: Physicality Score.
\vspace{-1.50em}
\end{table*}
\begin{figure}[H] 
\centering
\includegraphics[width=1.0\textwidth]{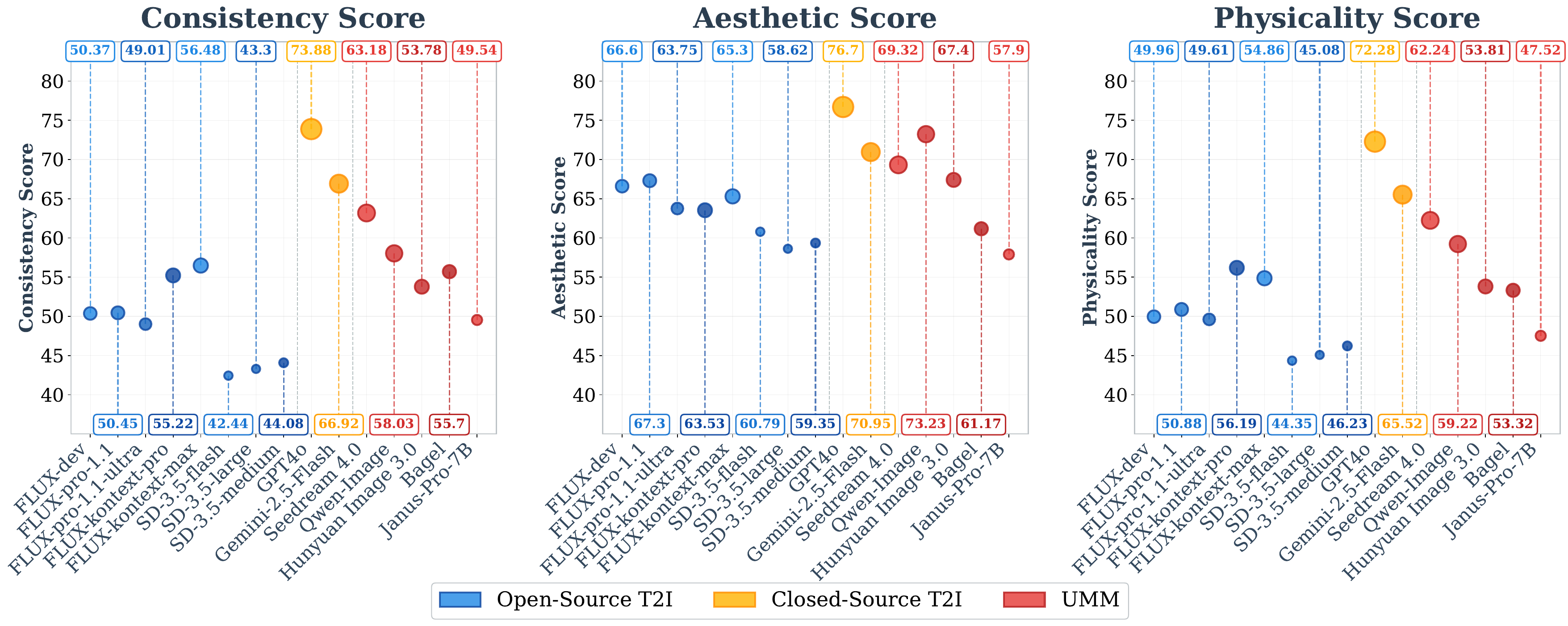}
\caption{Comparison of Model Performance on Key Evaluation Dimensions, shows the average scores of UMMs (Red), Open-Source T2I Models (Blue) and Closed-Source T2I Models (Yellow) across metrics.}
\label{fig:result_fig}
\vspace{-0.75em}
\end{figure}
\subsection{Evaluation Results}
The comprehensive evaluation of 15 models on the Envision benchmark, summarized in Table \ref{tab:comparison_average} and visualized in Figure \ref{fig:result_fig}, reveals a fundamental performance dichotomy between \textbf{UMMs}, \textbf{Open-Source T2I Models} and \textbf{Closed-Souced T2I Models}. Closed-Source T2I Models, exemplified by GPT-4o and Gemini-2.5-Flash-Image, establish dominant superiority across all evaluation axes, achieving peak scores in Consistency (73.88), Aesthetic (76.70), and Physicality (72.28). This advantage appears to stem from its massive data scale, along with scalable model sizes and proprietary training methods, endowing it with formidable generative capabilities. Open-Source T2I Models, represented by the FLUX and SD-3.5 series, demonstrate specialized proficiency in static visual fidelity, attaining high scores in Artistic Quality (FLUX-pro-1.1: 76.73). While excelling in texture rendering and stylistic composition, these models reveal a critical disconnect between surface-level aesthetics and underlying plausibility, as evidenced by substantial deficits in Factual Consistency (FLUX-kontext-pro: 60.33) and Physical Reliability (52.70). This fully reflects its lack of internalization of world knowledge. UMMs—architecturally unified models like Seedream and Qwen—occupy an intermediate position, effectively translating broad multimodal knowledge into enhanced scene understanding. This enables strong performance in knowledge-intensive domains such as Biology (Seedream: 76.92) and consistent advantages over Open-Source T2I Models in causal consistency-related metrics. Nevertheless, even these conceptually advanced UMMs currently trail behind state-of-the-art Closed-Source T2I Models in maximizing their potential for causal and physical reasoning. Most critically, the Spatial-Temporal Consistency dimension emerges as a universal challenge transcending all model categories, with the highest achieved score (GPT-4o: 67.42) remaining substantially limited—thereby validating the benchmark's core thesis that contemporary T2I generation models, despite their impressive static synthesis capabilities, maintain fundamentally gaps in coherent spatiotemporal reasoning.
\subsection{Key Insights from Envision}
\vspace{-0.75em}
\begin{figure}[H]
    \centering
    \includegraphics[width=1.0\textwidth]{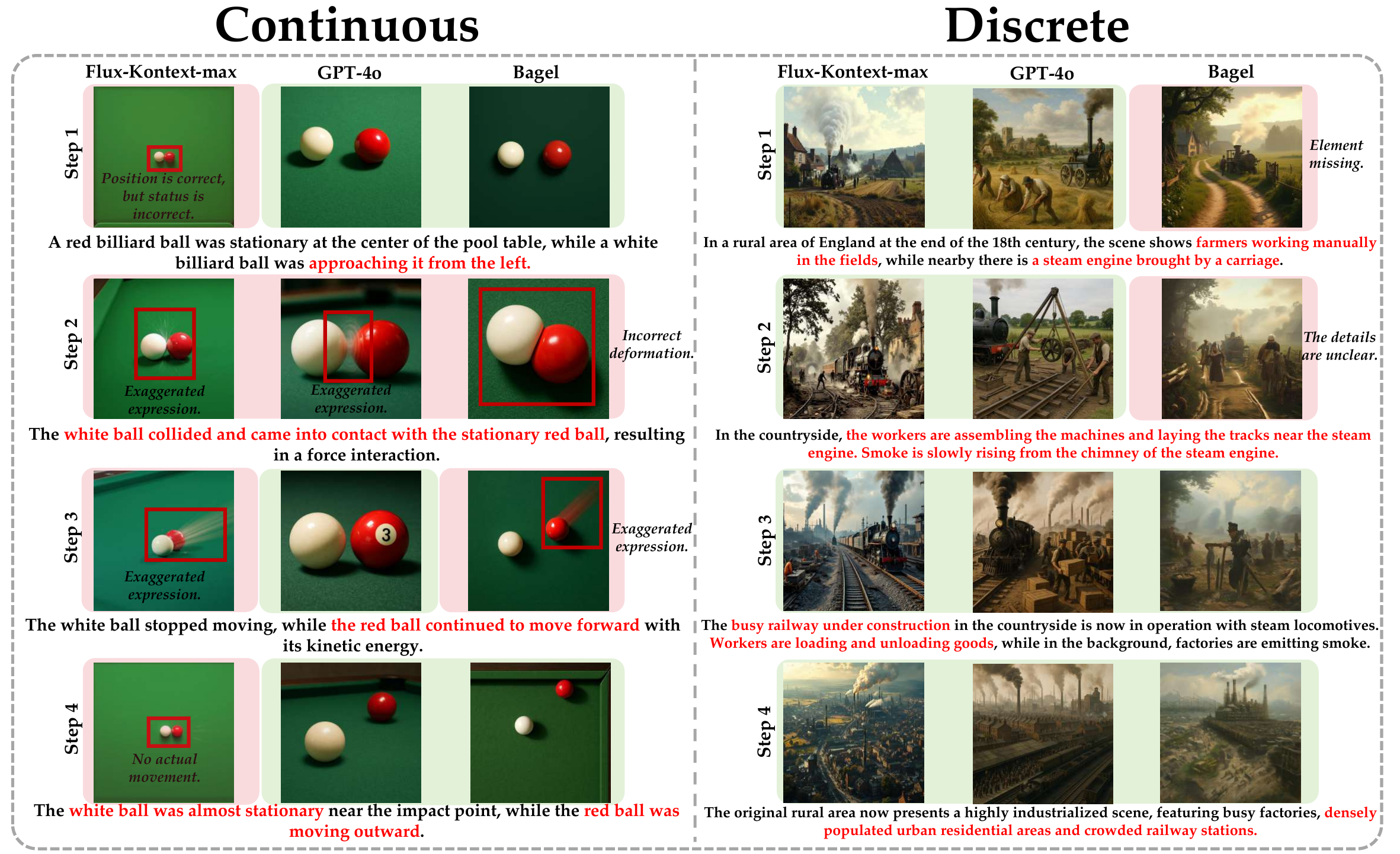}
    \caption{Visualization of Causal Event Progression and Failure Analysis. This figure compares the multi-step generative outputs of various generative model, including Flux-Kontext-max (Open-Source T2I Model), GPT-4o (Closed-Source T2I Model), Qwen-Image (UMM) and Bagel (UMM), across two distinct causal scenarios.}
    \label{fig:evidence}
    \vspace{-1em}
\end{figure}
\paragraph{Foundational Deficits in Dynamic Event Modeling.} The Envision benchmark reveals a fundamental limitation in contemporary multimodal T2I models: their inability to conceptualize and represent events as coherent spatio-temporal processes. Despite extensive training in large-scale static image datasets, these models demonstrate a systematic failure to internalize the causal mechanisms that govern dynamic evolution in the physical world. This deficit manifests as a striking performance dichotomy (as shown in Table \ref{tab:model_comparison_comprehensive} and Figure \ref{fig:evidence})---while models exhibit reasonable competence in handling commonsense scenarios that rely on static pattern recognition, they show profound deficiencies in scientific domains requiring strict adherence to physical laws and temporal constraints. The underlying challenge originates from the inherent tension between discrete and continuous event representation. Discrete event sequences can be partially addressed through static pattern matching and interpolation, whereas continuous processes demand deeply integrated world knowledge and explicit causal reasoning to maintain coherence across state transitions. Critically, our findings point to a breakdown in temporal reciprocal reasoning: the models fail to use the currently generated visual state as a validated memory scaffold to guide and constrain the subsequent causal evolution.
%This foundational gap highlights the intrinsic limitations of current architectural paradigms in capturing temporal dynamics and suggests that mere scaling of existing approaches may be insufficient to achieve genuine temporal understanding.
\vspace{-0.25em}
\paragraph{The Understanding-Generation Paradox: Fundamental Disconnect.} Our investigation establishes the Understanding-Generation Paradox as a critical failure of UMMs to execute effective reciprocal temporal reasoning. This paradox is not merely a performance gap but a systemic architectural conflict: the knowledge optimized for static, discrete representation (Understanding) is decoupled from the policy required for dynamic, continuous manipulation (Generation). Specifically, the generated visual state fails to serve as a rigorously validated causal memory trace for the next step, and the latent causal understanding fails to impose sufficient predictive constraints on the sequential generation policy. This breakdown is evidenced across two axes (Figure \ref{fig:evidence}): in discrete event scenarios, models exhibit \textbf{``visual fluency without causal fidelity,''} suggesting their success relies on statistical pattern interpolation without the underlying generative self-verification demanded by true causality. Conversely, in continuous process scenarios, models show ``nascent understanding with generative collapse,'' where the inability to maintain spatio-temporal consistency leads to the catastrophic breakdown of physical laws. Moreover, this divergence and the performance gap between UMMs and closed-source T2I models highlight a critical insight: the unification of understanding and generation remains significantly disjointed and conflicted. Resolving this paradox may require a paradigm shift toward explicitly demanding learning frameworks that interweave understanding and generation. This involves exploring more efficient and scalable architectures and training paradigms, ultimately validating and optimizing their internalization and representation of world knowledge through large-scale data.

\vspace{-0.25em}
\paragraph{Beyond Performance Gaps: Architectural and Conceptual Barriers.} 
The deficiencies revealed by the Envision are not limited to the shortcomings quantified by the surface Enscore but fundamentally expose a core issue in current UMMs: the disconnection between understanding and generation. The consistent inability of diverse architectures to enforce causal continuity and perform complex spatio-temporal reasoning suggests that these barriers stem not from architectural limitations but from a foundational conceptual flaw: the inductive bias primarily based on static, causally isolated individual images within the data and the training paradigm built upon it. When confronted with the need to simulate dynamic processes, the generation module often degrades into shallow pattern matching or semantic mapping, failing to effectively leverage the deeper, reasoned world knowledge potentially stored in the understanding module. Crucially, this limitation stems from the lack of architectural mechanisms that enforce the interleaving of understanding and generation, where semantic consistency and logical deductions explicitly constrain and guide the visual output across sequential steps. Conceptually, this requires incorporating video frame sequences or causally linked image sequences as native multi-image data modalities during training. This explicitly injects necessary spatio-temporal induction biases, ensuring that deeper world knowledge understanding is tightly interleaved with generation and genuinely manifested in multi-image visual narratives. To bridge the chasm between capturing isolated ``world states'' and simulating continuous ``world processes,'' future progress necessitates a fundamental shift from static pattern matching to integrated world models, even as the explicit Chain-of-Thought (CoT) mechanism serves as a promising, if external, universal bridge to enforce understanding's constraints and guide the multi-step causal generation process in the interim.

\section{Related Work}
\label{sec:related}
\subsection{Unified Multimodal Models (UMMs)}
Unified Multimodal Models (UMMs) aim to integrate diverse tasks like visual understanding and generation within a single architecture, promoting cross-modal synergy and reducing complexity. Recent works fall into several paradigms: \textbf{Autoregressive (AR) Models} like Chameleon~\cite{team2024chameleon}, Janus~\cite{wu2025janus}, and Emu3~\cite{wang2024emu3} tokenize visual inputs for sequential generation, while Show-O~\cite{xie2024showo} incorporates a discrete-diffusion schedule for refined token prediction. \textbf{AR with Diffusion Head} approaches, such as Transfusion, often keep a pre-trained MLLM frozen and route its features to an external generator~\cite{tong2024metamorph, shi2024lmfusion} to enable complex multimodal interactions. Alternatively, \textbf{Integrated Transformers}~\cite{zhao2024monoformer, chen2024diffusion} unify different paradigms in one backbone to eliminate bottlenecks. To enhance scalability, the \textbf{Mixture-of-Transformers (MoT)} paradigm~\cite{liang2024mixture, deng2025emerging} introduces a sparse, modular design, as seen in Bagel. However, progress in model architecture must be gauged through rigorous evaluation. Current evaluations, however, still rely heavily on static single-image benchmarks, which can mask fundamental limitations like the disconnection and opposition between understanding and generation.

\subsection{Multimodal Understanding and
Generation Benchmarks}
Landscape of evaluation benchmarks for multimodal T2I model is evolving, shifting from isolated assessments of singular capabilities toward more integrated, comprehensive evaluations. In the domain of understanding, the research focus has moved from basic perception to an ``thinking-with-
images'' paradigm that incorporates generative abilities\cite{yue2024mmmu, su2025twi}. Concurrently, generative evaluation has progressed beyond foundational metrics of semantic fidelity\cite{jiang2024genai}. It now involves calibrating the performance of unified models against traditional generative counterparts in text-to-image\cite{ramesh2021zerot2i} and image editing\cite{ye2025imgedit}, specifically targeting capability gaps across key dimensions such as consistency\cite{wang2025pref, huang2023t2icomp, huang2024t2ifactual}, world knowledge\cite{niu2025wise, li2025science, fu2024commonsense, meng2024phybench, chow2025physbench}. While there is a burgeoning effort to illustrate the benefits of unified interaction through reasoning tasks\cite{sun2025t2ireason} and the analysis of coupled versus decoupled generation-understanding frameworks\cite{xie2025mme, zou2025unimmmu,li2025unieval,ghazanfari2025chain_frame,huang2025vchain,zhang2025chain_focus}, a significant limitation persists: these benchmarks remain predominantly anchored to single-image generation tasks, which inherently fail to capture the causal and temporal dynamics of event processes. We address this limitation by introducing multi-image sequential scenarios, which transition evaluation from a conventional ``static" paradigm to a ``dynamic". By meticulously decomposing and reconstructing previous evaluation metrics, Envision enables a more intuitive, authentic, and multifaceted assessment of a model's actual performance across all established dimensions.

\section{Conclusion}
\label{sec:conclusion}
In conclusion, Envision establishes that prevailing paradigms in multimodal T2I model, anchored in static single-image generation and evaluation, are fundamentally insufficient for achieving genuine world understanding. The Envision benchmark and its findings expose a critical ``Understanding-Generation Paradox'', revealing that proficiency in depicting isolated scenes does not equate to an internalized model of dynamic, causal processes. The pronounced performance gap between UMMs and T2I models, coupled with the universal bottleneck in spatiotemporal reasoning, underscores a systemic limitation of current architectures, data and training regimens. Our results compellingly argue that future progress hinges on a foundational shift—from optimizing for static pattern matching to architecting models capable of world simulation. 

\clearpage
\newpage
\bibliographystyle{plainnat}
\setcitestyle{numbers}
\bibliography{paper}

\clearpage
\newpage
\beginappendix

\section{More Details about Envision}
\label{sec:details}
This appendix serves as a supplementary compendium to the main text, providing exhaustive methodological details, formal definitions, and extended analyses that underpin the Envision.

\section{Definition of Categories in Envision}
\label{appendix:category_definition}
This appendix provides a formal definition of each category and sub-category within the benchmark, elucidating the specific cognitive capabilities they are designed to probe.

\subsection{Natural Science}
This category of evaluation models assesses the internalized understanding of fundamental laws governing nature. Achieving success requires not only maintaining robust semantic consistency under spatiotemporal constraints, but also internalizing world knowledge and deducing the sequential natural scientific processes—including physical, chemical, and biological phenomena—that compose each event frame and the entire multi-image sequence of events within a complete event progression.

\begin{itemize}
    \item \textbf{Physics:} This sub-category evaluates qualitative and semi-quantitative reasoning about core principles from classical mechanics, thermodynamics, optics, and electromagnetism. Scenarios are engineered to have unambiguous, visually verifiable outcomes based on physical laws. The model must demonstrate an understanding of \textit{state transitions} governed by forces, energy, and conservation laws.\\
    \textbf{Exemplar Task:} As shown in Figure \ref{fig:domain}. ``A white billiard ball rolls across a table and strikes a stationary red billiard ball. Show the sequence of what happens during and after the collision.'' A correct sequence must depict the transfer of momentum between the balls, with the white ball slowing down or stopping while the red ball moves away, thereby illustrating the principles of conservation of momentum and energy in elastic collisions.

    \item \textbf{Chemistry:} Tasks in this domain probe the model's comprehension of molecular-level interactions and their macroscopic consequences. This includes reasoning about reaction kinetics (precipitation, combustion), stoichiometry, and phase transitions. The benchmark requires models to infer the visual outcomes of chemical processes, moving beyond symbolic representations to dynamic visualizations of transformation.\\
    \textbf{Exemplar Task:} As shown in Figure \ref{fig:domain}. ``Clear lead nitrate solution and potassium iodide solution are mixed together in a beaker. Show the sequence of what happens immediately after mixing.'' A correct sequence must depict the instantaneous formation of a bright yellow precipitate of lead iodide, demonstrating a double displacement precipitation reaction and the transition from soluble reactants to insoluble products.

    \item \textbf{Biology:} Evaluations focus on quintessential biological processes that operate across various scales, from individual to ecological. This includes modeling life cycles (\textit{e.g.,} metamorphosis), species evolution, predator-prey population dynamics, and ecosystem succession. The model must reason about temporal progressions driven by biological imperatives like growth, reproduction, natural selection, and resource competition.\\
    \textbf{Exemplar Task:} As shown in Figure \ref{fig:domain}. ``A whale carcass sinks to the deep ocean floor. Show the sequence of its decomposition over time.'' A coherent sequence should illustrate the ecological succession: from initial consumption by mobile scavengers like sharks, through colonization by smaller organisms on the remaining tissues and bones, to the final stage where specialized bone-eating worms and bacteria break down the skeleton, completing the nutrient cycle.

    \item \textbf{Geography:} This sub-category focuses on long-term geomorphological processes and the spatial relationships between physical and human features on Earth's surface. It challenges models to extrapolate the slow, yet deterministic, evolution of landscapes and human-environment interactions. Reasoning is based on causal drivers like tectonic forces, erosion, deposition, and human modification.\\
    \textbf{Exemplar Task:} As shown in Figure \ref{fig:domain}. ``An island volcano erupts. Show the sequence from the eruption to the ecological recovery over an extended period.'' A coherent sequence should visually demonstrate the causal stages: (1) the initial volcanic eruption with lava flows and ash covering the landscape; (2) the gradual cooling and solidification of volcanic materials, forming new landforms; (3) the pioneering stage where initial plant life and organisms colonize the barren terrain; (4) the eventual establishment of a recovered ecosystem with diverse flora and fauna, illustrating long-term ecological succession following a major disturbance.
    
    \item \textbf{Meteorology:} This sub-category focuses on short-to-medium-term atmospheric processes and weather phenomena. It evaluates a model's ability to reason about the formation, progression, and dissipation of weather systems based on thermodynamic principles and fluid dynamics. Tasks require understanding the visual manifestations of atmospheric states and their sequential evolution.\\
    \textbf{Exemplar Task:} As shown in Figure \ref{fig:domain}. ``Over a Gobi desert landscape, show the sequence from the formation of rain clouds to the end of a thunderstorm.'' A correct sequence must depict the causal stages: (1) the initial formation and gathering of cumulus clouds in the sky; (2) the development into dark, towering cumulonimbus clouds; (3) the occurrence of rainfall and potential lightning over the arid land; (4) the final dissipation of clouds and the clearing of the sky, leaving a moistened ground.
    
\end{itemize}

\subsection{Cultural and History}
This category aims to evaluate a model's alignment with shared human knowledge, social conventions, and historical narratives. It assesses the model's comprehension of \textit{intent, cultural logic}, and \textit{social causality}, and deduces the core semantic alignment components within multi-image narrative processes alongside other relevant variations across the entire pictorial narrative—elements that are often implicit and context-dependent.

\begin{itemize}
    \item \textbf{Cultural \& History:} This unified area probes a model's knowledge of stereotypical human activities and their evolution. At the micro-level, this involves understanding the script-like sequences of everyday events. At the macro-level, it requires modeling the impact of pivotal historical developments on material culture, social organization, and daily life. The core challenge is to ground abstract historical narratives or social conventions in their concrete, visual manifestations across time.\\
    \textbf{Exemplar Task:} As shown in Figure \ref{fig:domain}. ``Show the founding and early growth of Apple Computer in a garage during the 1970s.'' A coherent sequence must visualize the key stages: (1) the initial scene of a suburban garage with two young entrepreneurs surrounded by electronic components and early blueprints; (2) the development phase with assembled circuit boards and intensified prototyping activities; (3) the small-scale production stage with multiple finished units and packaging materials; (4) the transition to a recognized company, symbolized by the garage's eventual transformation into a historic landmark.
\end{itemize}

\section{Prompt Structure}
\label{subsec:prompt_structure}

To ensure the generation of causally coherent and visually plausible multi-image sequences for evaluation, we designed a structured JSON-based prompt template. This template has been manually designed and validated, clearly defining the sequence of phases while maintaining strict narrative constraints between each stage. Additionally, we developed specialized prompt styles to facilitate multi-step prompt combinations and their corresponding explanations.

\paragraph{JSON Schema Definition}
Each event process scenario is represented as a JSON array containing four sequential steps. The schema is defined as follows:

\begin{verbatim}
{
  "prompts": [
    {
      "step": 1,
      "prompt": "",
      "explanation": ""
    },
    {
      "step": 2,
      "prompt": "",
      "explanation": ""
    },
    {
      "step": 3,
      "prompt": "",
      "explanation": ""
    },
    {
      "step": 4,
      "prompt": "",
      "explanation": ""
    }
  ]
}
\end{verbatim}

\subsection{Event-Level Prompt Consistency Constraints}
The prompt structure enforces several critical constraints to ensure evaluation validity:
\begin{itemize}
    \item \textbf{Narrative Logic of Events:} For the prompt design of each step, we remind that \textit{``this is the Xth event frame in a four-stage event progression, ...''} and the explanation for that stage provides a detailed description and hints for the events occurring in the prompt of that step. Ultimately, the prompts of the four steps serve as four event frames, forming a multi-image event sequence.
    \item \textbf{Viewpoint Consistency:} All four frames must maintain a reasonable camera position, angle, and field of view relative to the scene being depicted, unless a change in perspective is explicitly motivated by the narrative. This ensures that the model can focus on the dynamic evolution of the event itself, rather than being distracted by inconsistent visual framing. It is crucial for establishing a coherent spatial context and enabling accurate comparison of states across different temporal stages, thereby facilitating the assessment of a model's understanding of spatiotemporal relationships.
    \item \textbf{Environmental Stability:} Throughout the entire process, particularly in natural science scenarios such as physics and chemistry, lighting conditions, background elements, and experimental apparatus must remain consistent to maintain narrative coherence. For other scenarios, emphasis is placed on lighting, scene composition, and the arrangement of elements within the scene.
    \item \textbf{Temporal Progression:} Each step must represent a logical, causally-driven, and progressive stage within the entire event process. The sequence should visually articulate a clear ``before-and-after'' relationship between consecutive frames, demonstrating a plausible evolution of states rather than a mere substitution of scenes. This requires the model to internalize not just individual states, but the dynamic mechanisms—that govern the transitions between them. 
\end{itemize}

\subsection{Event-Level Prompt Framework}
The four-stage multi-image sequence follows a carefully designed causal narrative framework:

\begin{enumerate}
    \item \textbf{Step 1 - Initial State:} This frame establishes the narrative's baseline, introducing the core entities, environment, and all initial conditions. A well-defined initial state provides the necessary foundation from which the entire causal chain can unfold, anchoring the sequence in a specific and stable starting point.
    \item \textbf{Step 2 - Early Interaction:} This frame depicts the inciting incident or the initial catalyst that disrupts the initial state. It captures the first dynamic interaction, a key decision by an agent, or the onset of a natural process. This step is critical for initiating the causal chain, showing the immediate consequences of the interaction and setting the narrative in motion.
    \item \textbf{Step 3 - Progressive Transformation:} This frame represents the core development phase of the event. It visualizes the significant intermediate state where the initial interactions lead to pronounced changes, cascading effects, or complex developments. This step tests the model's ability to simulate the ongoing process, maintain logical state transitions, and depict the evolving relationships between entities as the narrative progresses toward its culmination.
    \item \textbf{Step 4 - Final Resolution:} This frame concludes the event sequence by showcasing the ultimate outcome and steady state resulting from the preceding transformations. It demonstrates the final configuration of the system, the long-term consequences of the actions, and the fulfillment of the causal chain. A coherent resolution should align with established principles, providing a logically and visually satisfying endpoint to the narrative.
\end{enumerate}

This structured approach ensures that the generated sequences maintain subject plausibility while providing a standardized basis for evaluating model performance on T2MI tasks, with the JSON format further facilitating automated processing and analysis of evaluation results across diverse data domains. It is noteworthy that, whereas early CLIP-based models constrained prompt tokens to a limit of 77, our current focus has shifted toward advanced T2I models and UMMs; accordingly, the original prompts have been appropriately enriched with descriptive details—guided by expert annotation and supplemented by GPT-4o—to enhance expressiveness without compromising fairness, with each prompt strictly limited to within 100 tokens.

\section{Quadripartite Event Frame Structure}
\label{app:event-structure}

The Envision benchmark employs a standardized four-frame sequence to represent a complete causal event. This specific structural choice is grounded in a deliberate trade-off between representational expressivity, computational feasibility, and evaluative rigor.

\subsection{Multiple images in Visual Narrative Generation}

Visual narrative generation exists on a spectrum defined by temporal granularity. At one extreme lies \textbf{single-image generation (T2I)}, which captures a solitary, static moment. This paradigm is inherently limited for evaluating dynamic processes, as it lacks temporal extension and cannot express state transitions. At the other extreme lies \textbf{video generation (T2V)}, which models continuous, high-frame-rate sequences. While powerful, T2V introduces significant computational complexity and often focuses on short-term motion dynamics rather than high-level, causal event progression.

The \textbf{multi-image sequence (T2MI)}, and specifically the four-frame structure, positions itself as an optimal intermediate representation:

\begin{itemize}
    \item \textbf{Sufficiently Discrete to Avoid Complexity:} A sequence of four images is computationally less intensive to generate and evaluate than a high-frame-rate image sequence or video. It avoids the context window limitations and prohibitive memory costs associated with processing long visual sequences in current transformer-based or diffusion-based models.
    
    \item \textbf{Sufficiently Continuous to Model Causality:} Crucially, four frames provide the minimal yet sufficient number of distinct states to define a complete \textbf{causal narrative arc}. As established in our prompt structure, these frames map directly onto the canonical stages of a narrative: \textit{Initial State (S1), Inciting Interaction (S2), Progressive Transformation (S3), and Final Resolution (S4)}. This structure forces the model to conceptualize an event not as a collection of independent scenes, but as a chain of causally linked states, thereby discouraging mere associative pattern matching and promoting genuine causal reasoning.
\end{itemize}

\subsection{Cognitive and Evaluative Advantages}

The four-frame paradigm is aligned with cognitive theories of event perception, where humans naturally segment continuous experience into discrete, meaningful units bounded by causal boundaries.

From an evaluation perspective, this structure provides several critical advantages:

\begin{enumerate}
    \item \textbf{Amplification of Consistency Errors:} A sequence of four frames provides multiple transition points (S1$\rightarrow$S2, S2$\rightarrow$S3, S3$\rightarrow$S4) where inconsistencies in object attributes, spatial relationships, and temporal logic can be detected. A three-frame sequence might be too concise to reveal progressive errors, while a longer sequence (\textit{e.g.,} five or six frames) might introduce redundant states without providing proportional diagnostic value, thereby increasing evaluation cost without a commensurate gain in insight.
    
    \item \textbf{Mitigation of Contextual Interference:} When a long sequence of prompts or images is input into a model as a single context, there is a risk of ``context dilution'' or ``forgetting,'' where early context influences the generation of later frames in an uncontrolled manner. The four-frame sequence is short enough to fit comfortably within the context windows of modern large models, ensuring that all prompts are considered concurrently during generation. 
    
    \item \textbf{Structured Diagnostic Granularity:} The four-stage arc allows for precise fault localization. A failure in S2 indicates a model's inability to initiate a causal chain from a stable state. A breakdown between S3 and S4 suggests a failure in projecting a process to its logical conclusion. This granularity is essential for providing actionable insights into model limitations.
\end{enumerate}

\section{Scoring Criteria of Metrics}
\label{sec:scoring_criteria}

To ensure a rigorous and nuanced evaluation of model performance on the Envision benchmark, we have established a comprehensive scoring framework. This framework is built upon three core dimensions—\textbf{Consistency}, \textbf{Physicality}, and \textbf{Aesthetics}—each decomposed into specific, actionable sub-dimensions. Each sub-dimension is scored on a discrete 0-5 scale, where 0 represents a catastrophic failure and 5 represents flawless execution. The following sections detail the rationale and scoring rubrics for each dimension.

\subsection{Consistency Dimension}

\begin{figure*}[t!]
\centering
\includegraphics[width=\textwidth]{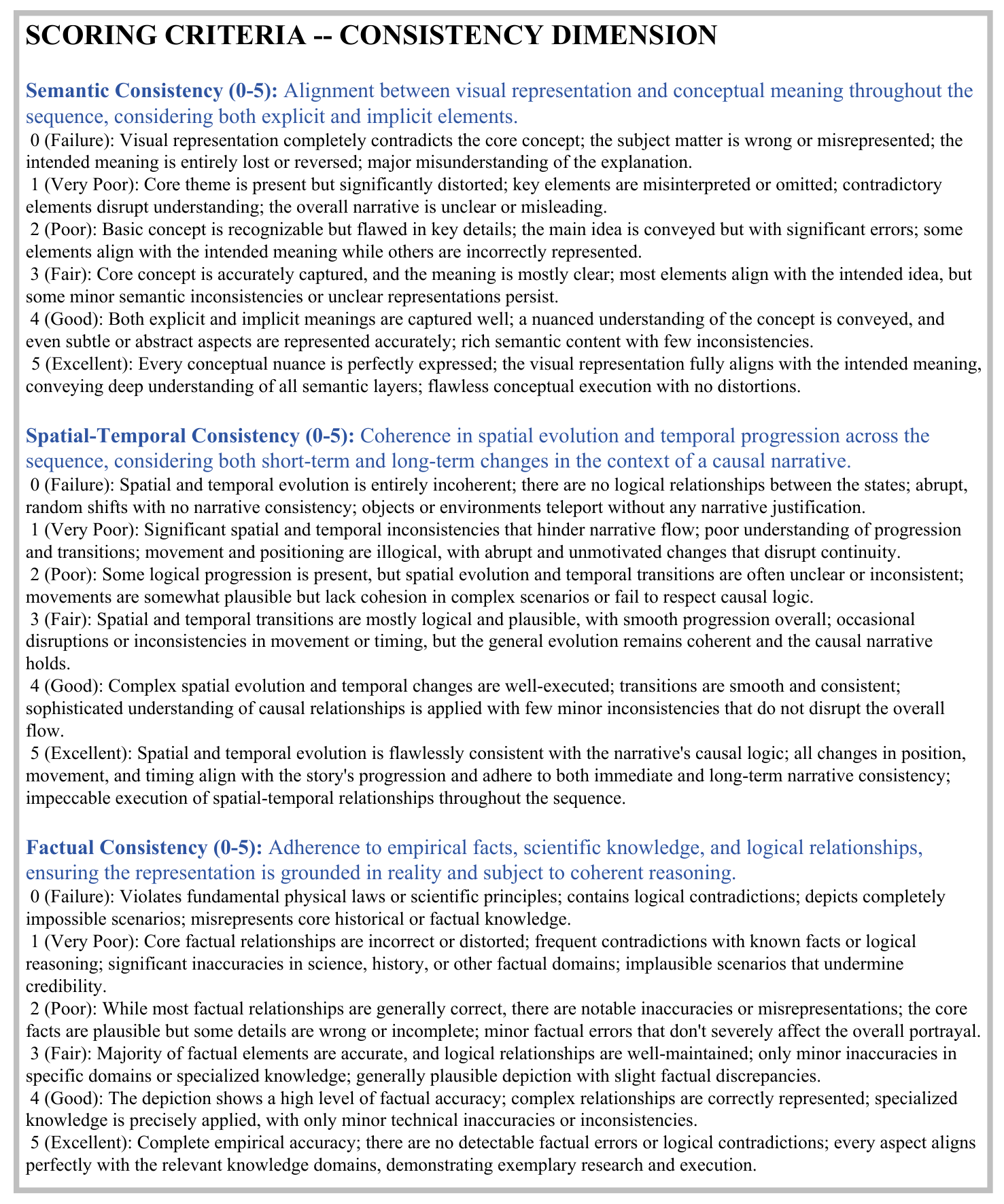}
\caption{Scoring Criteria for the \textbf{Consistency} dimension, evaluating the spatiotemporal, semantic, and factual consistency of generated multi-image sequences. Each is scored on a 0--5 scale, where 0 indicates catastrophic failure and 5 represents flawless execution.}
\label{fig:score_con}
\end{figure*}

The Consistency dimension evaluates the logical, semantic, and factual consistency of the generated sequence. It ensures that the narrative is stable, meaningful, and grounded in reality (see Figure \ref{fig:score_con}).

\begin{itemize}
    \item \textbf{Spatio-Temporal Consistency:} Evaluates the coherence of spatial relationships and their logical evolution over time across the image sequence. It ensures that objects, characters, and the environment maintain a stable presence, follow plausible motion trajectories, and undergo changes in a continuous and causally sound manner. A high score (4-5) requires smooth, logical transitions where all spatial changes (\textit{e.g.,} movement, deformation) are physically plausible and temporally coherent, adhering to a consistent narrative timeline. A low score (0-1) indicates severe discontinuities, such as objects teleporting, erratically changing size or shape, backgrounds inconsistently shifting, or a complete breakdown in the logical progression of events.
    \item \textbf{Semantic Consistency:} Assesses the alignment between the visual output and the conceptual meaning of the prompt. It judges whether the generated images correctly convey the intended story, action, or concept. A high score (4-5) demands that both explicit and implicit meanings are captured, whereas a low score (0-1) signifies a fundamental misunderstanding or contradiction of the core theme.
    \item \textbf{Factual Consistency:} Gauges the adherence to empirical facts and established logical relationships. This sub-dimension is critical for scenarios involving scientific, historical, or commonsense knowledge. A top score of 5 indicates complete empirical accuracy with no detectable errors, while a score of 0 reflects violations of fundamental facts or physical laws.
\end{itemize}

\subsection{Aesthetic Dimension}

\begin{figure*}[t!]
\centering
\includegraphics[width=\textwidth]{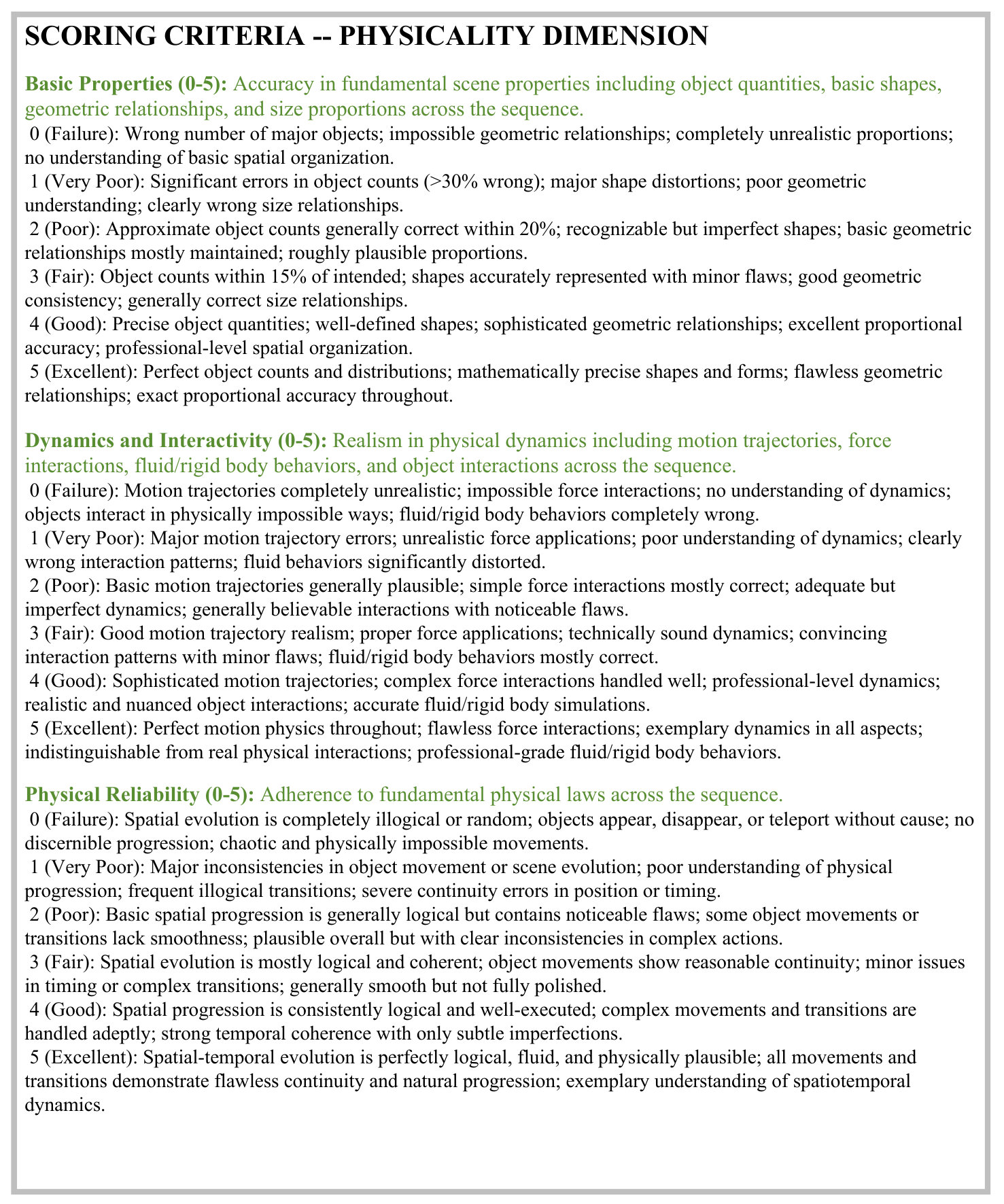}
\caption{Scoring Criteria for the \textbf{Physicality} dimension, assessing the model's adherence to physical laws and plausibility of dynamic processes. Scores range from 0 (failure) to 5 (excellent).}
\label{fig:score_a}
\end{figure*}

While secondary to causal reasoning, the Aesthetic dimension remains essential for evaluating the overall quality and usability of the generated content. It ensures that the sequences are not only correct but also visually compelling (see Figure \ref{fig:score_a}).

\begin{itemize}
    \item \textbf{Expressivenes:} Evaluates the effectiveness and emotional impact of the visual storytelling. It assesses how well the composition, character expressions, lighting, and color are used to convey the narrative's mood, action, and underlying causality. A high score (4-5) indicates a dynamic and evocative sequence that powerfully communicates the event's drama and progression, while a low score (0-1) reflects a flat, inert, or emotionally disconnected portrayal that fails to engage the viewer in the narrative.
    \item \textbf{Aesthetic Quality:} Assesses the overall artistic merit, style coherence, and emotional impact. It judges whether the visual style is appropriate, consistent, and effectively supports the narrative through masterful execution of composition, color, and lighting. A high score (4-5) indicates a visually stunning and stylistically harmonious sequence that deeply enhances the storytelling. A low score (0-1) reflects a poorly executed, inconsistent, or aesthetically unappealing result that fails to engage the viewer artistically.
    \item \textbf{Authenticity:} Measures the believability and naturalness of the generated imagery, judging how closely it approximates the intended reality (whether photorealistic or stylistically consistent). A high score (4-5) signifies that the sequence is visually convincing and free of jarring artificial artifacts, making it indistinguishable from a real photograph or a masterfully consistent artistic rendering. A low score (0-1) is given for obvious synthetic flaws, unrealistic textures, unnatural lighting, or a cartoonish appearance that breaks immersion and undermines plausibility.
\end{itemize}

\subsection{Physicality Dimension}

\begin{figure*}[t!]
\centering
\includegraphics[width=\textwidth]{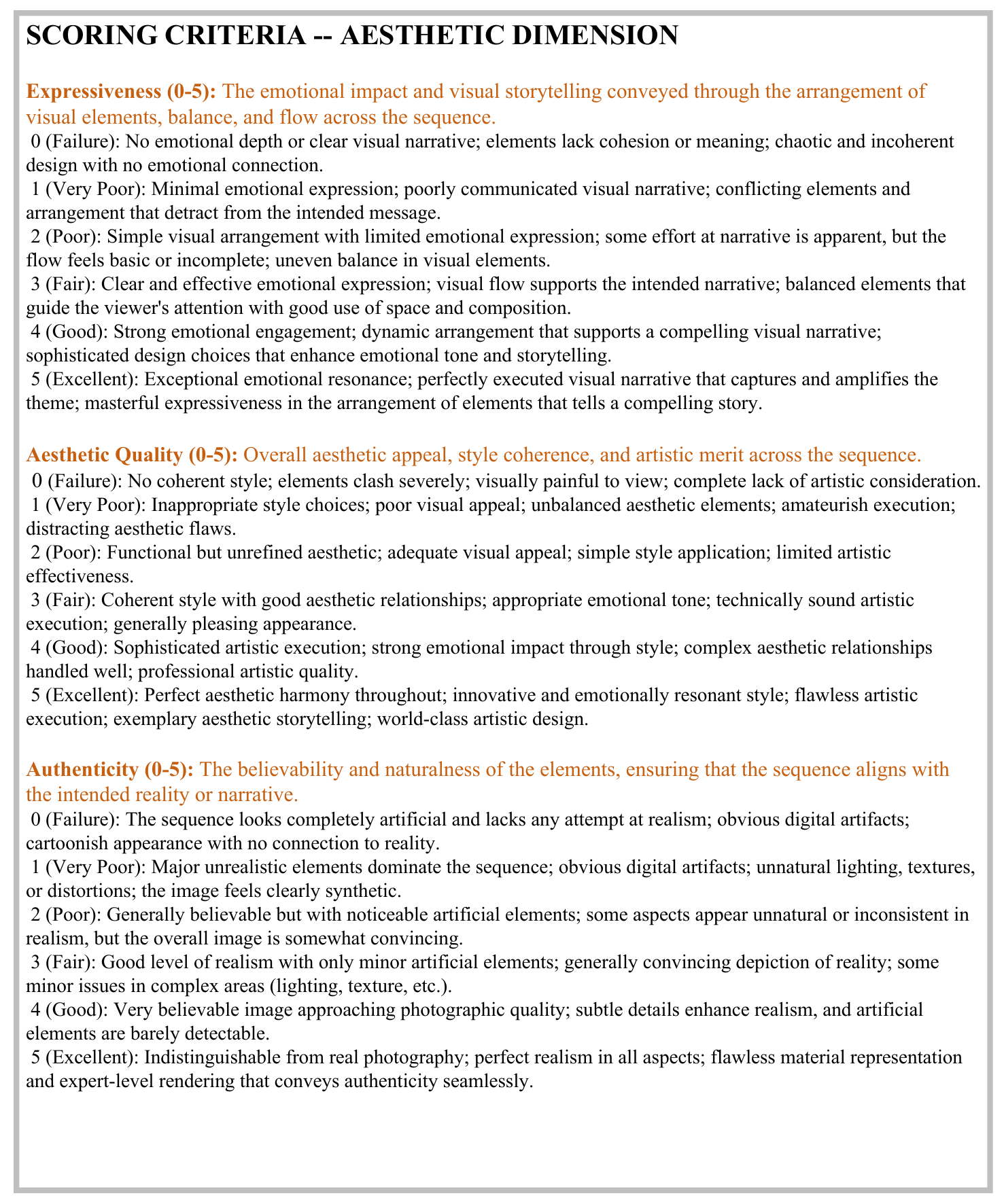}
\caption{Scoring Criteria for the \textbf{Aesthetics} dimension, capturing the visual quality and narrative expressiveness of generated sequences. Scores range from 0 (failure) to 5 (excellent).}
\label{fig:score_phy}
\end{figure*}

The Physicality dimension is the cornerstone of our causal reasoning evaluation. It scrutinizes the model's adherence to the principles of the physical world, from macroscopic geometry to microscopic material properties (see Figure \ref{fig:score_phy}).

\begin{itemize}
    \item \textbf{Basic Properties:} Evaluates the accuracy and stability of fundamental scene attributes throughout the sequence. This includes the precise maintenance of object counts, the preservation of fundamental geometric shapes and sizes, and the consistency of scale and proportions between objects. A high score (4-5) requires that all entities retain their core attributes without spurious appearance, disappearance, or morphing, while a low score (0-1) indicates severe errors such as objects vanishing, replicating, or exhibiting impossible and unstable shapes.
    \item \textbf{Dynamics \& Interactivit:} Assesses the plausibility of physical movements and interactions between entities. This dimension scrutinizes the realism of motion trajectories, force transmissions, collision outcomes, and the behavior of materials (\textit{e.g.,} fluids, rigid bodies). A high score (4-5) depicts interactions that are governed by coherent physics, such as a falling object following a parabolic arc or a collision resulting in momentum transfer, whereas a low score (0-1) reveals physically impossible motions, such as objects passing through each other or moving without applied force.
    \item \textbf{Physical Reliability:} Gauges the adherence to fundamental physical laws and principles across the entire sequence. This serves as the ultimate test of the model's internalized world model, checking for violations of thermodynamics, conservation laws, gravity, and other invariant principles. A high score (4-5) demonstrates a sequence that is entirely plausible within our physical reality, while a low score (0-1) is assigned for catastrophic failures, such as energy being created from nothing, perpetual motion, or blatant defiance of gravity.
\end{itemize}

\subsection{Scoring Rubric and Aggregation}

Each of the aforementioned sub-dimensions is scored by human experts or a qualified large vision-language model (VLM) using a standardized rubric:

\begin{itemize}
    \item \textbf{5 (Excellent):} Flawless execution regarding the specific sub-dimension. No detectable errors.
    \item \textbf{4 (Good):} Minor, inconsequential deviations. The core intent and logic are fully preserved.
    \item \textbf{3 (Fair):} Noticeable errors that do not fundamentally undermine the causal chain or primary objective.
    \item \textbf{2 (Poor):} Significant errors that violate causal narrative or visual consistency.
    \item \textbf{1 (Very Poor):} Fundamentally flawed, bearing little resemblance to a correct outcome.
    \item \textbf{0 (Failure):} Irrelevant, nonsensical, or non-compliant output.
\end{itemize}

To emphasize the importance of causal reasoning, the final \textbf{Envision (Overall) Score} is calculated as a weighted sum of the primary dimension scores: \textbf{40\%} for Physicality, \textbf{40\%} for Consistency, and \textbf{20\%} for Aesthetics. This formulation ensures that a model's ability to reason about event processes is the dominant factor in its final evaluation.

\section{VLM Score Reliability Validation}

The Envision-Score metric, designed to evaluate T2I models and UMMs on their ability to generate temporally consistent, physically plausible, and aesthetically coherent multi-image sequences, requires rigorous validation to ensure its reliability. This section outlines the validation process used to assess the consistency and robustness of the scores produced by the scoring framework, as well as the mechanisms that underpin the trustworthiness of these evaluations.

\subsection{Multi-Trial Evaluation}

To ensure statistical validity, we conducted \textbf{K independent evaluation trials} for each generated sequence, maintaining consistent conditions across all trials. For each sequence, comprising four steps image in a event progression, the scoring process was repeated multiple times (K=5), providing a comprehensive data set that allows for a detailed reliability analysis. (see the Figure \ref{fig:overall_score})

Each trial resulted in an evaluation tuple \( T_k = (s_k, r_k) \), where \( s_k \in [0, 5]^n \)
represents the fine-grained score vector across all nine sub-dimensions, and \(r_k \) denotes the textual rationale provided by the evaluator model, explaining the score assignment for each sub-dimension.

This multi-trial method ensures that the scoring process is robust to any inconsistencies that might arise from random variations in evaluation conditions, model interpretation, or inherent ambiguities in the generated sequences.

\subsection{Quantitative Stability Analysis}

For each sub-dimension score, the \textbf{upper and lower bounds} \( s_{d,i}^{\text{max}}, s_{d,i}^{\text{min}}  \) and \textbf{standard deviation} \( \sigma_{sd,i} \) were computed across the K trials. This step provides a quantitative measure of the stability and variance of the scores, reflecting the reliability of the evaluation process. The lower the standard deviation, the more consistent the model's performance is, indicating a high level of reliability in the scoring framework.

The standard deviation as:

\begin{equation}
\sigma_{sd,i} = \sqrt{\frac{1}{K} \sum_{k=1}^{K} (s_{k,d,i} - \mu_{sd,i})^2}
\end{equation}

The upper and lower bounds are defined as:
\begin{align}
s_{d,i}^{\text{max}} &= \max( s_{k,d,i} : k = 1, \ldots, K ) \\
s_{d,i}^{\text{min}} &= \min( s_{k,d,i} : k = 1, \ldots, K )
\end{align}
By analyzing the distribution of scores across multi-trials, we can assess whether the scoring metric reliably reflects the quality of model outputs or whether variability in evaluation might be due to inconsistencies in the generated sequences or the evaluation model itself.

\subsection{Qualitative Analysis for Evaluation}

\begin{figure*}[htbp]
    \centering
    
    % First row: Quality metrics
    \begin{subfigure}{0.23\textwidth}
        \centering
        \includegraphics[width=\linewidth]{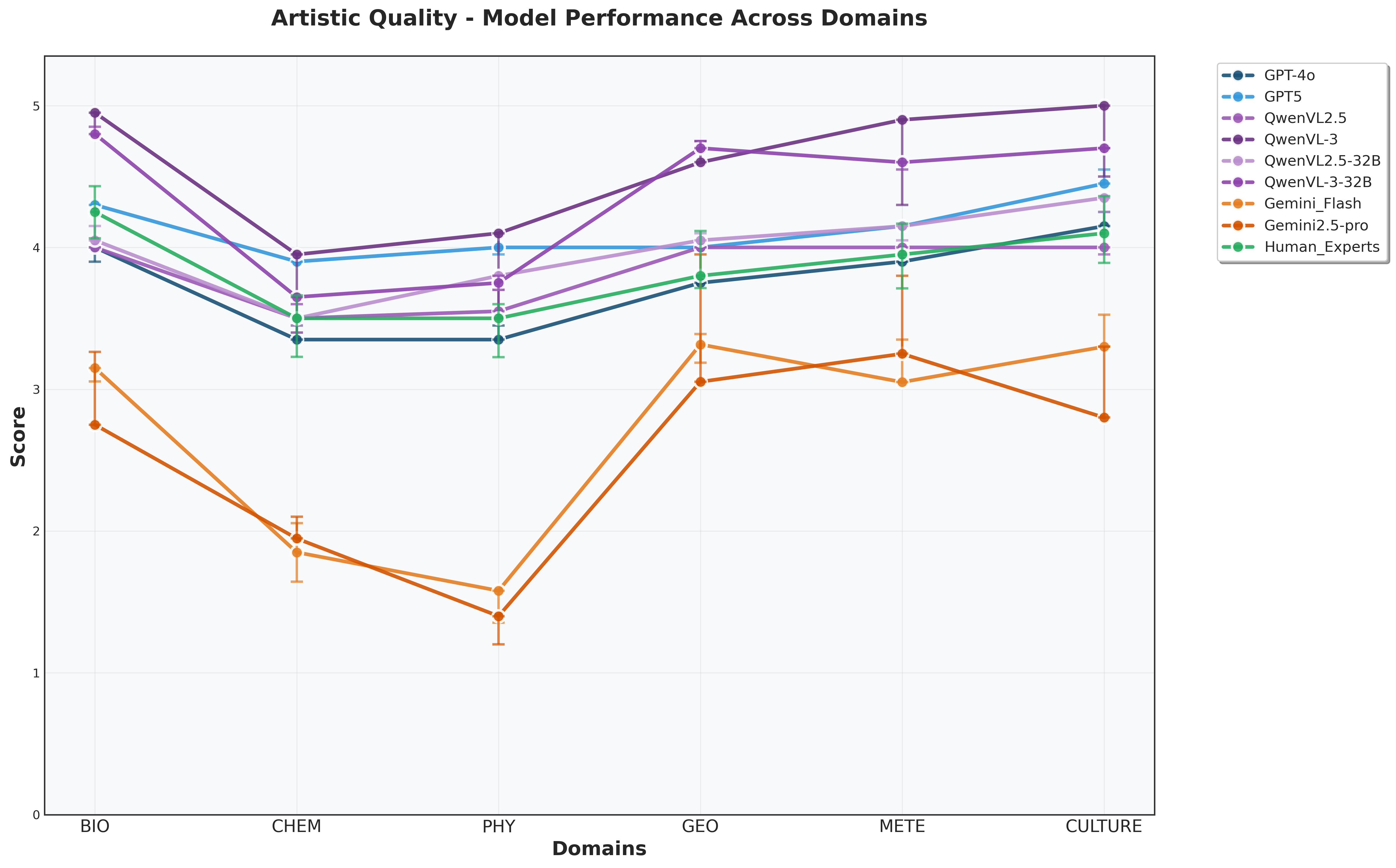}
        \caption{AQ}
        \label{fig:artistic_quality}
    \end{subfigure}
    \hfill
    \begin{subfigure}{0.23\textwidth}
        \centering
        \includegraphics[width=\linewidth]{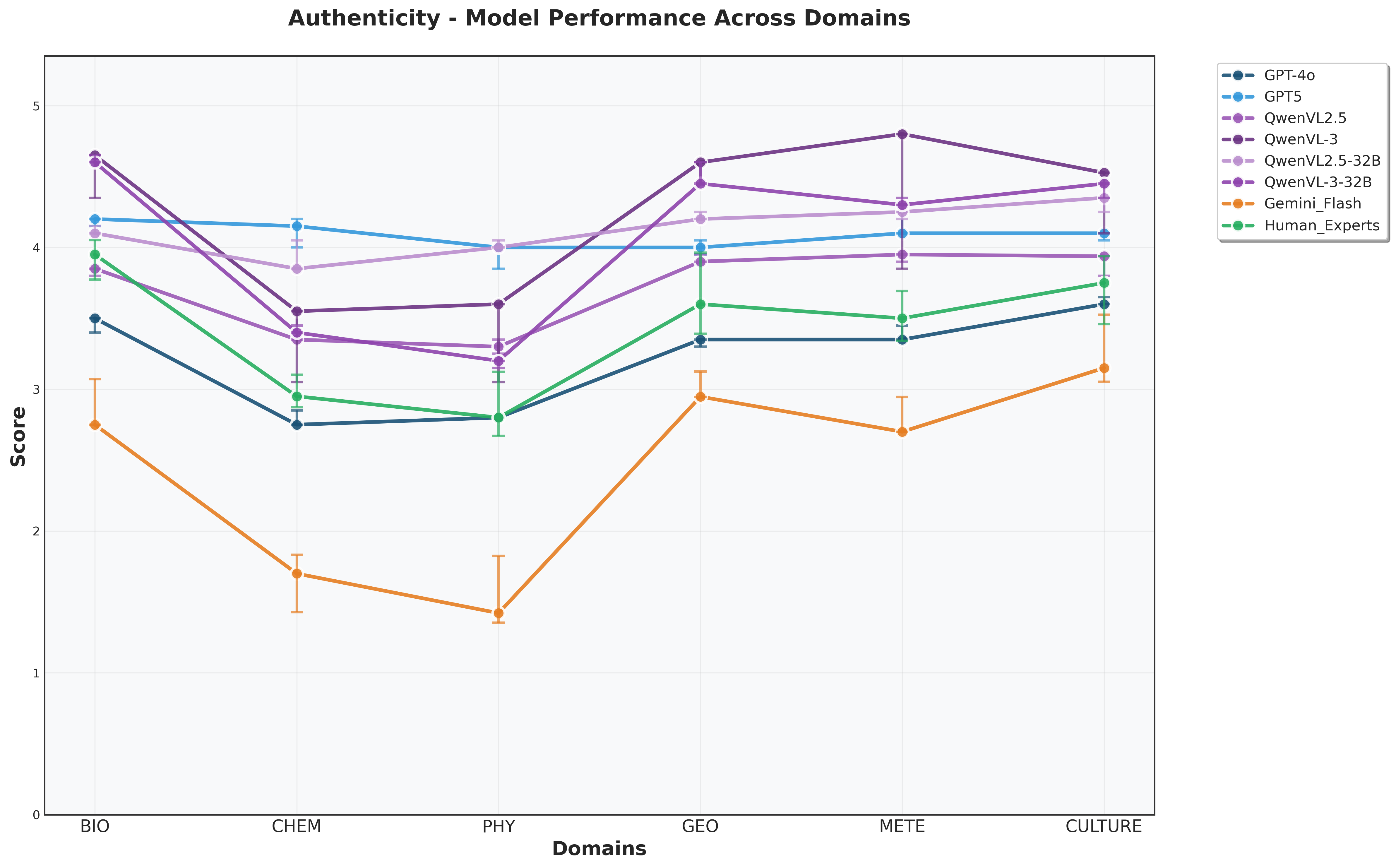}
        \caption{Auth}
        \label{fig:authenticity}
    \end{subfigure}
    \hfill
    \begin{subfigure}{0.23\textwidth}
        \centering
        \includegraphics[width=\linewidth]{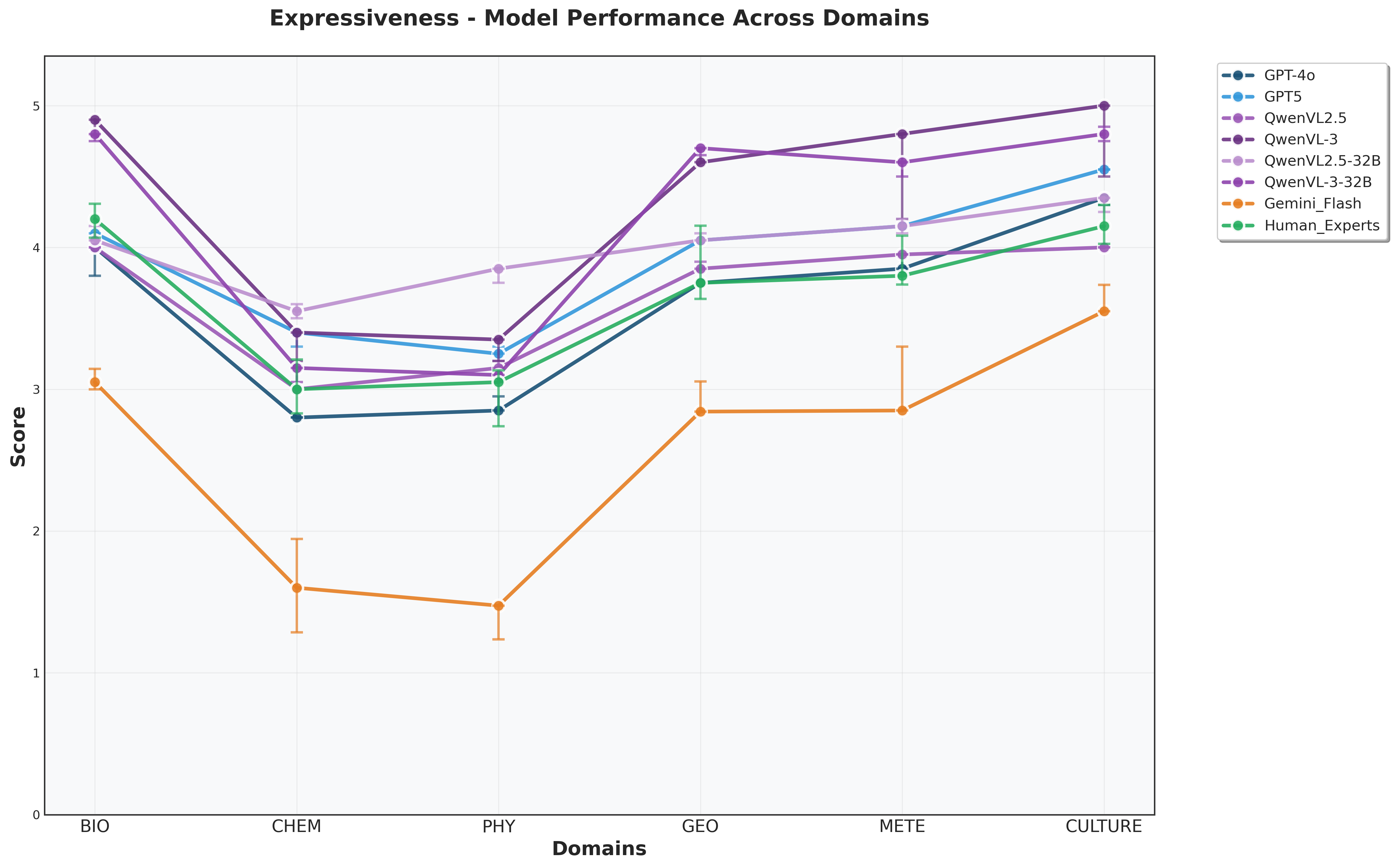}
        \caption{Exp}
        \label{fig:expressiveness}
    \end{subfigure}
    \hfill
    \begin{subfigure}{0.23\textwidth}
        \centering
        \includegraphics[width=\linewidth]{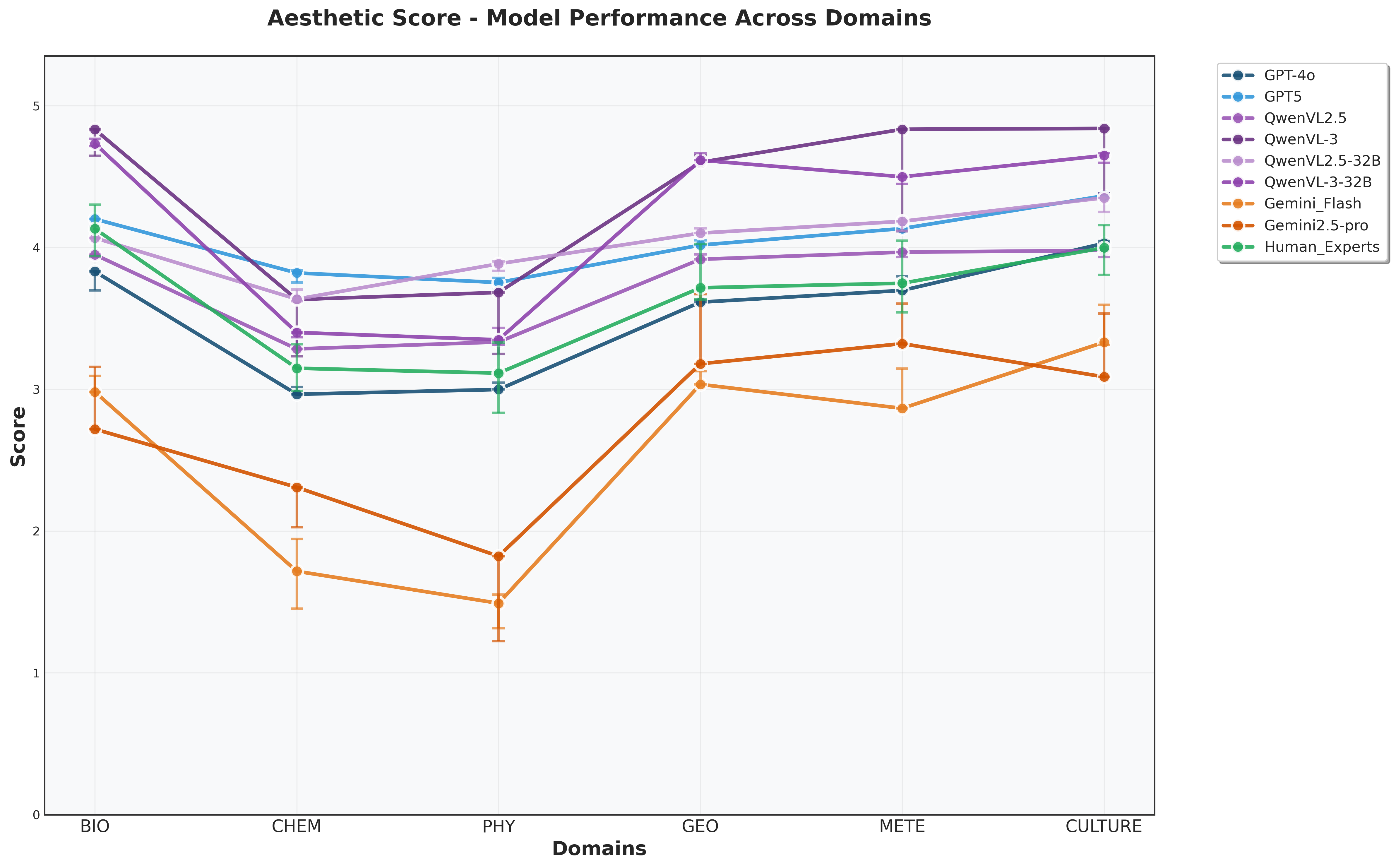}
        \caption{AS}
        \label{fig:aesthetic_score}
    \end{subfigure}
    
    % Second row: Consistency metrics
    \vspace{0.2cm}
    \begin{subfigure}{0.23\textwidth}
        \centering
        \includegraphics[width=\linewidth]{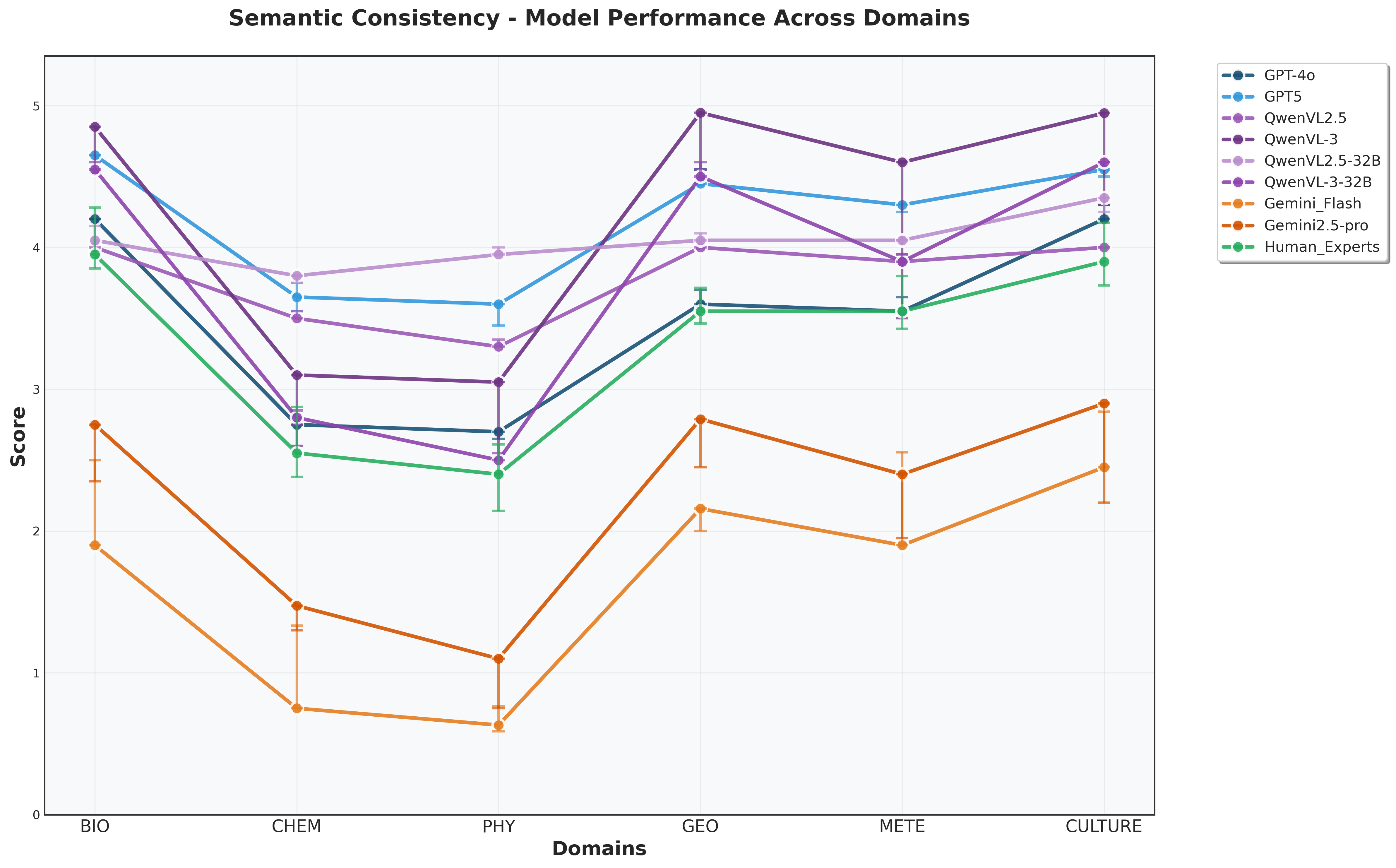}
        \caption{SC}
        \label{fig:semantic_consistency}
    \end{subfigure}
    \hfill
    \begin{subfigure}{0.23\textwidth}
        \centering
        \includegraphics[width=\linewidth]{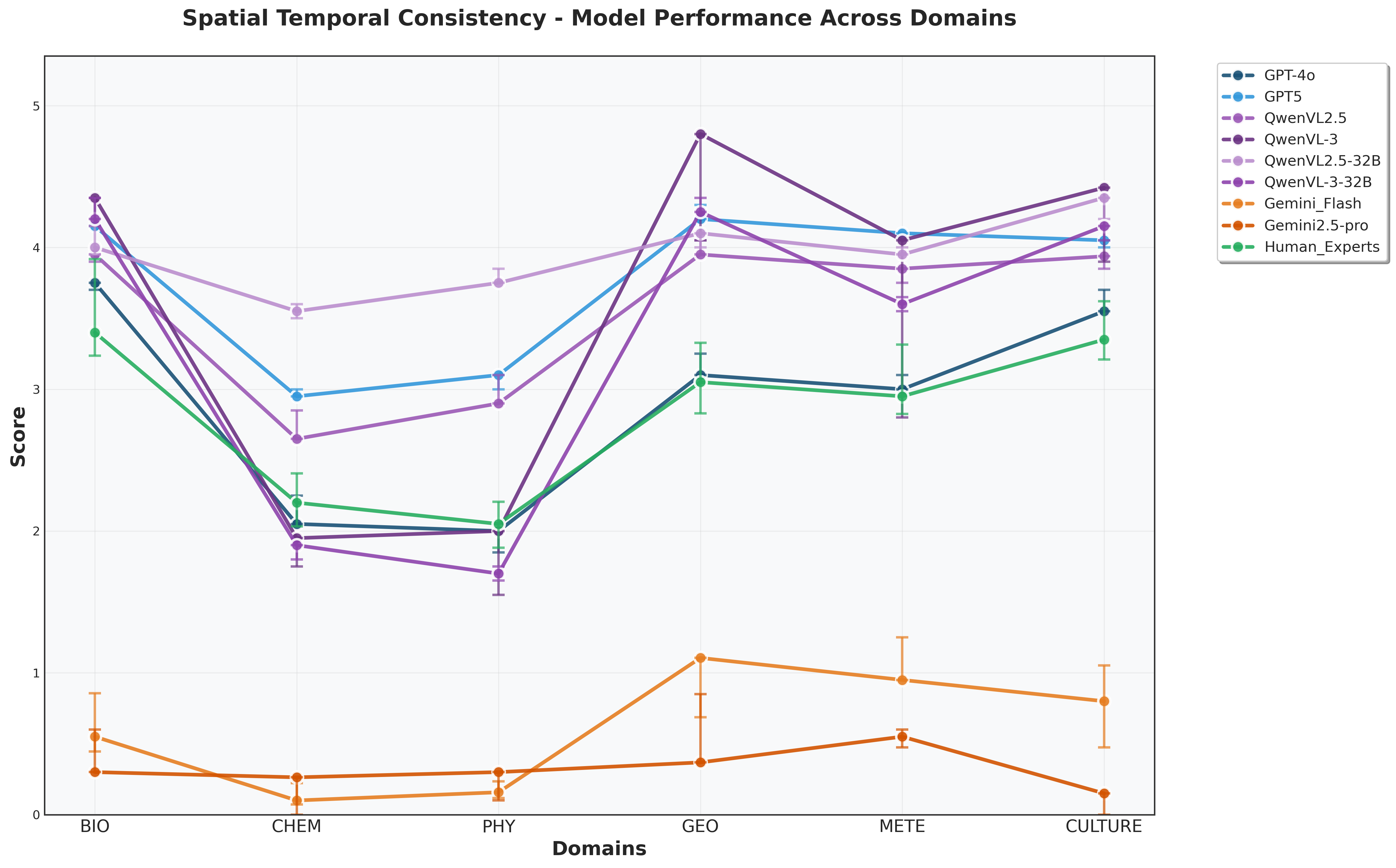}
        \caption{STC}
        \label{fig:spatial_temporal_consistency}
    \end{subfigure}
    \hfill
    \begin{subfigure}{0.23\textwidth}
        \centering
        \includegraphics[width=\linewidth]{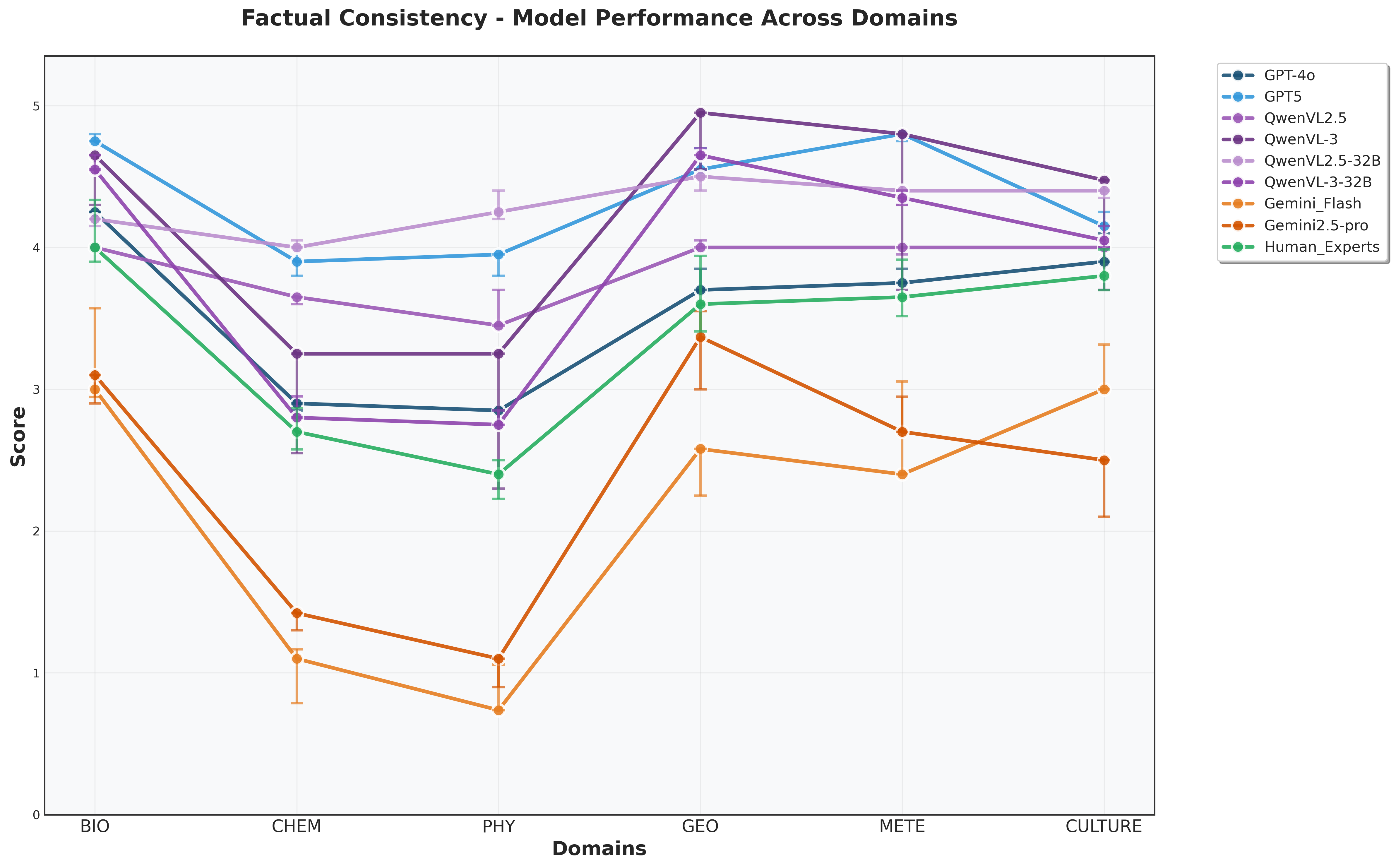}
        \caption{FC}
        \label{fig:factual_consistency}
    \end{subfigure}
    \hfill
    \begin{subfigure}{0.23\textwidth}
        \centering
        \includegraphics[width=\linewidth]{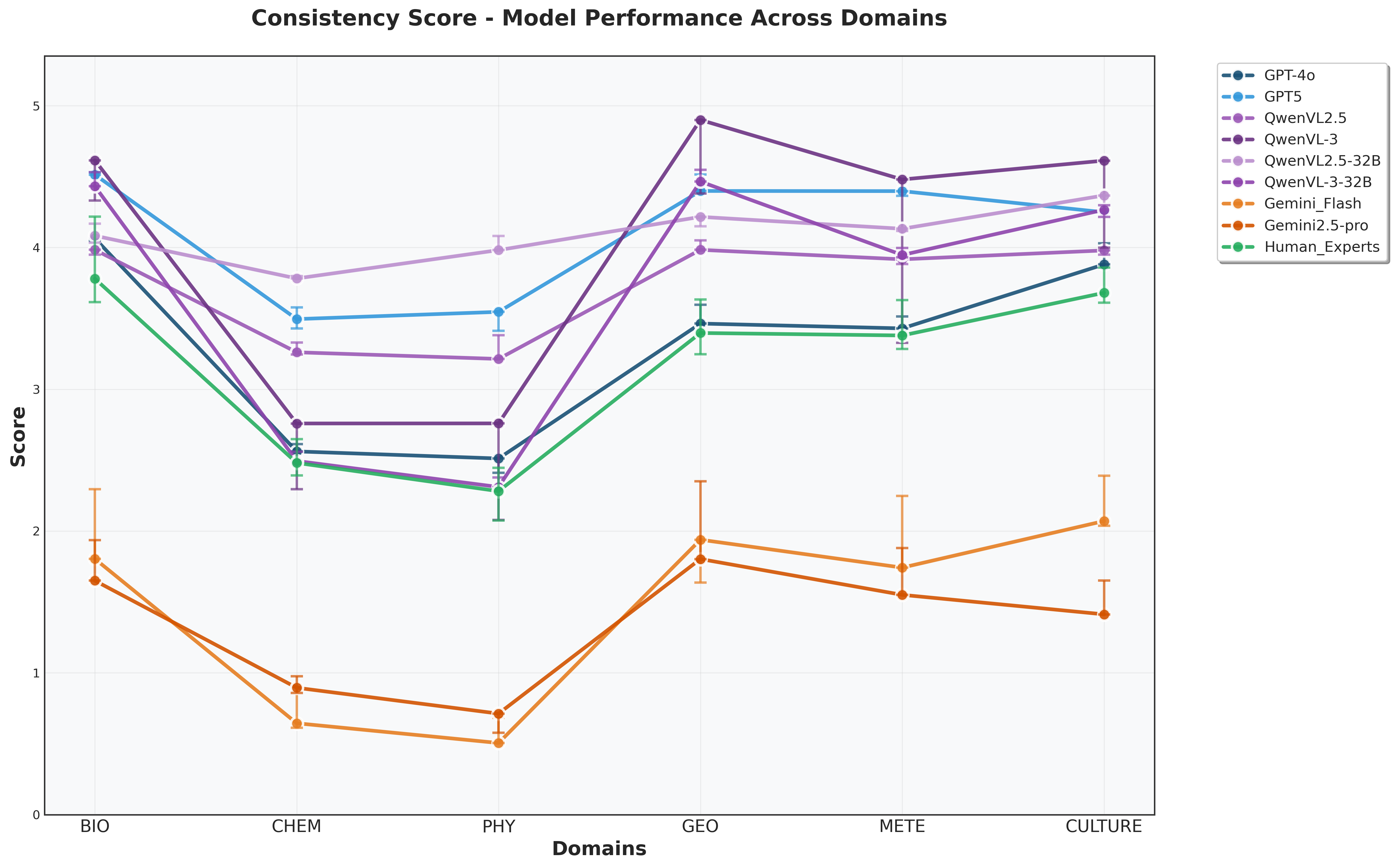}
        \caption{CS}
        \label{fig:consistency_score}
    \end{subfigure}
    
    % Third row: Physicality metrics
    \vspace{0.2cm}
    \begin{subfigure}{0.23\textwidth}
        \centering
        \includegraphics[width=\linewidth]{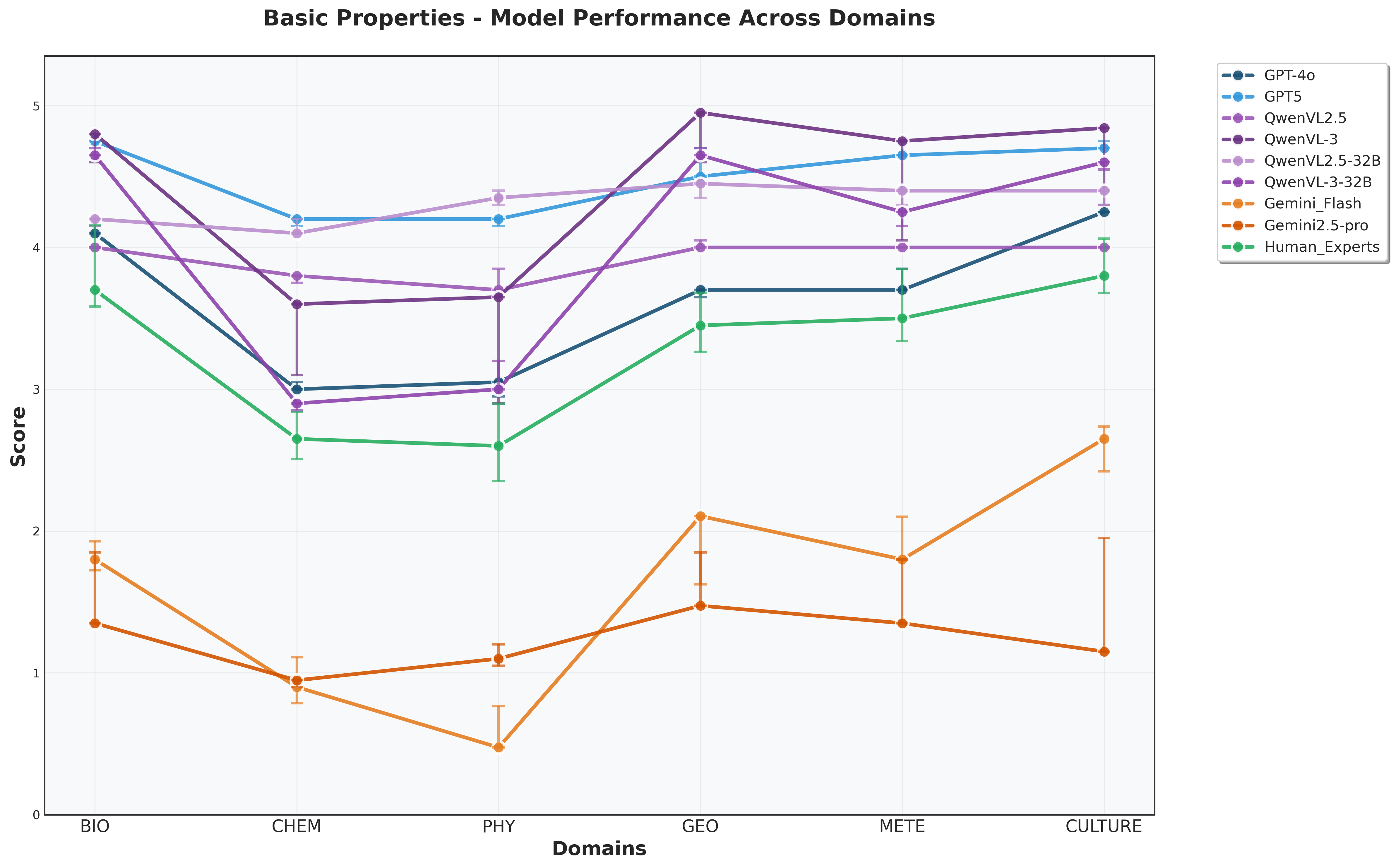}
        \caption{BP}
        \label{fig:basic_properties}
    \end{subfigure}
    \hfill
    \begin{subfigure}{0.23\textwidth}
        \centering
        \includegraphics[width=\linewidth]{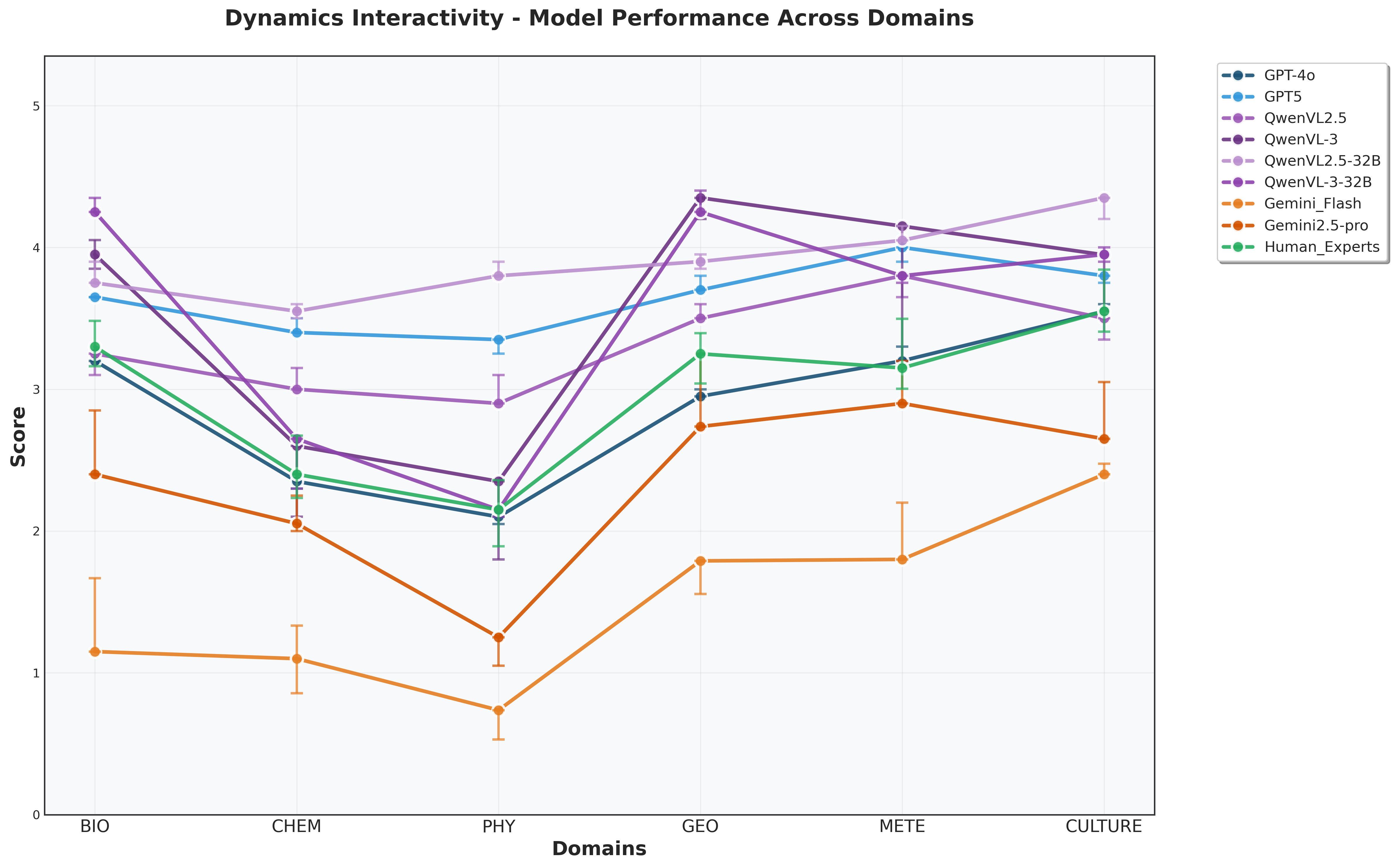}
        \caption{DI}
        \label{fig:dynamics_interactivity}
    \end{subfigure}
    \hfill
    \begin{subfigure}{0.23\textwidth}
        \centering
        \includegraphics[width=\linewidth]{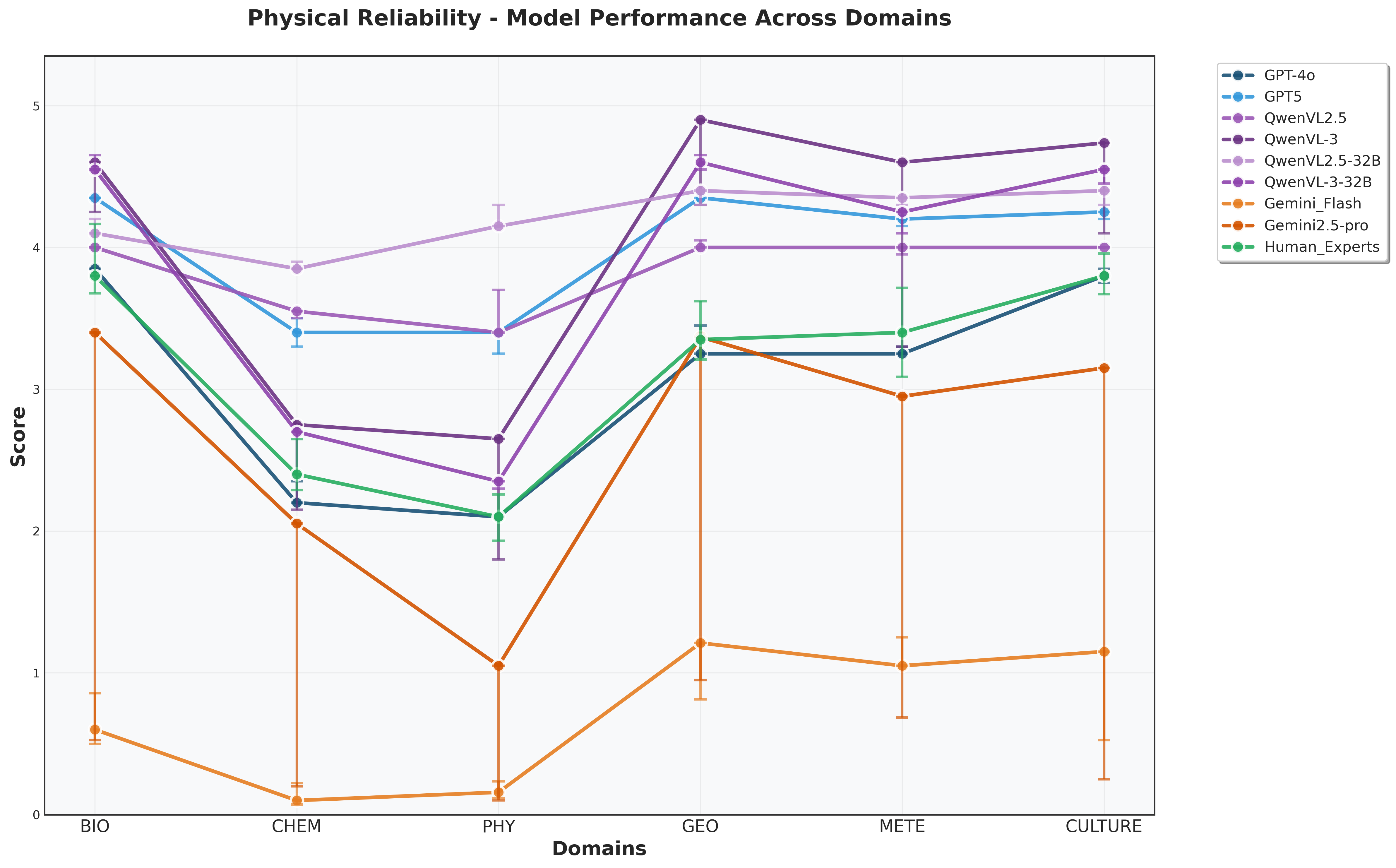}
        \caption{PR}
        \label{fig:physical_reliability}
    \end{subfigure}
    \hfill
    \begin{subfigure}{0.23\textwidth}
        \centering
        \includegraphics[width=\linewidth]{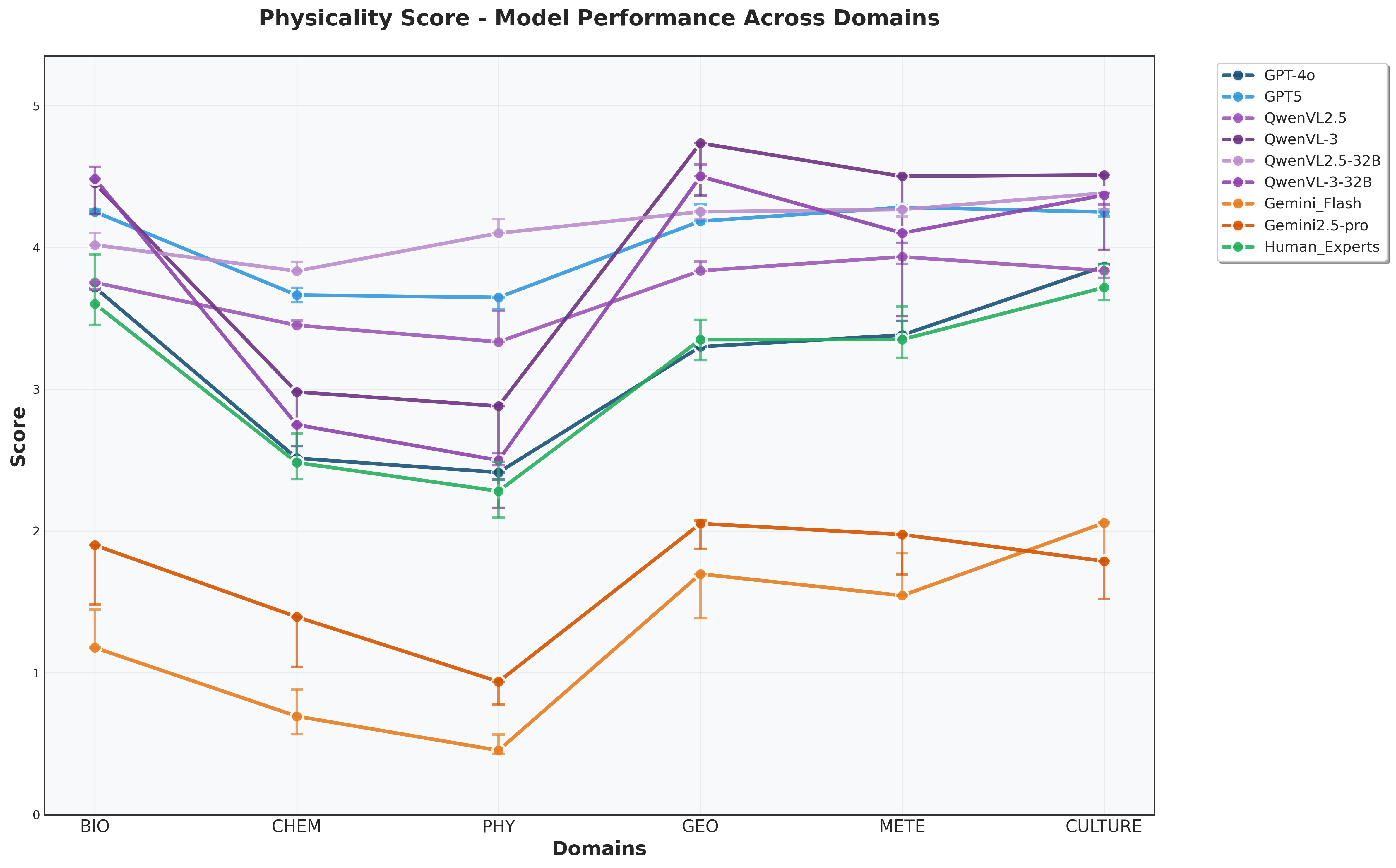}
        \caption{PS}
        \label{fig:physicality_score}
    \end{subfigure}
    
    % Fourth row: Overall score (centered)
    \vspace{0.2cm}
    \centering
    \hspace{2cm}
    \begin{subfigure}{0.8\textwidth}
        \centering
        \includegraphics[width=\linewidth]{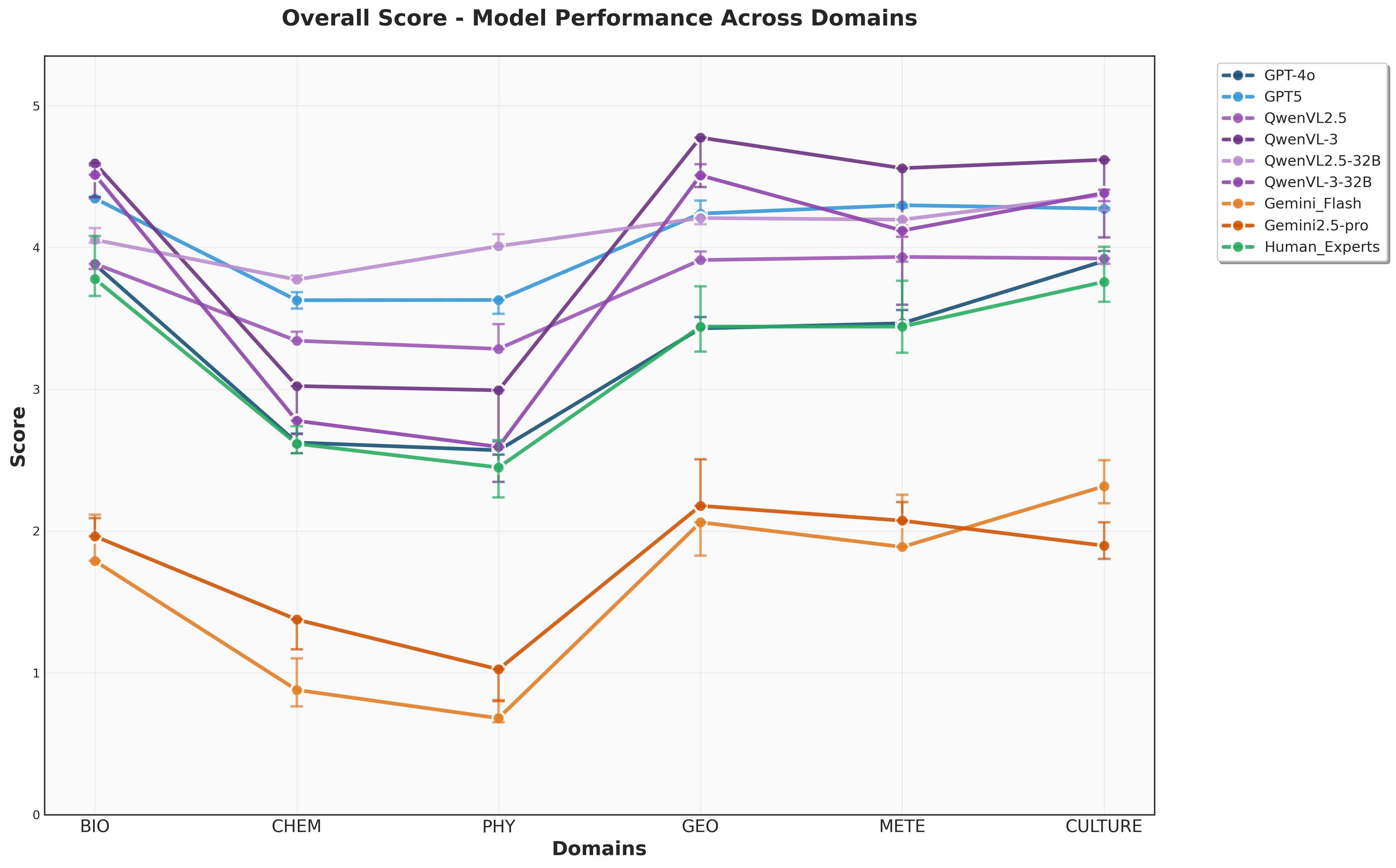}
        \caption{Overall}
        \label{fig:overall_score}
    \end{subfigure}
    
    \caption{Comprehensive evaluation of model performance across multiple dimensions: (a-d) aesthetic metrics (AQ, Auth, Exp, AS); (e-h) consistency metrics (SC, STC, FC, CS); (i-l) physicality metrics (BP, DI, PR, PS); (m) overall performance score.}
    \label{fig:model_performance_comprehensive}
\end{figure*}

To complement the quantitative analysis, a \textbf{qualitative consistency analysis} was conducted with the involvement of five \textbf{domain experts} (PhD-level evaluators) who assessed 50 GPT-4o-generated multi-image sequences per domain (as shown in Figure \ref{fig:model_performance_comprehensive}). This step provided a human-driven evaluation of the coherence of the generated sequences, focusing on how well they aligned with real-world causal processes and spatiotemporal dynamics.

These human judgments were compared to the scores assigned by GPT-4o, the automated evaluator. The results showed a strong agreement between human experts and the GPT-4o assessments, demonstrating that the model’s scoring methodology is not only reliable but also consistent with the expert-level judgment. Any discrepancies in scores were systematically reviewed, contributing to refining the scoring algorithm and ensuring that the evaluations capture the true causal coherence of the sequences.

\subsection{Selection of GPT-4o as Final Evaluator}

Based on the results of the qualitative and quantitative analyses, \textbf{GPT-4o} was chosen as the final automated evaluator. This decision was driven by its superior alignment with expert judgments and its ability to produce semantically stable patterns across multiple evaluation dimensions. By minimizing random variance in scoring and providing consistent rationales for its scores, GPT-4o was confirmed to be the most reliable tool for evaluating the causal and physical consistency of the generated sequences.

\newpage
\begin{table*}[t]
\centering
\caption{Comparative Analysis of Text-to-Visual Generation Modalities: Requirements and Benchmarks.}
\label{tab:modality_comparison_compact}
\footnotesize
\renewcommand{\arraystretch}{1.4}
\setlength{\tabcolsep}{10pt}
\begin{tabular}{@{}l>{\raggedright\arraybackslash}p{7cm}>{\raggedright\arraybackslash}p{4.5cm}@{}}
\toprule
\textbf{Modality} & \textbf{Core Requirements} & \textbf{Evaluation Benchmarks} \\
\midrule

\parbox[t]{2.2cm}{\textbf{T2I}\\{\footnotesize\textit{(Text-to-Image)}}} & 
\underline{Single-frame Synthesis}
\begin{itemize}
  \item \textbf{Visual Fidelity:} High-resolution, artifact-free image generation, image aesthetic quality.
  \item \textbf{Semantic Consistency:} Precise matching between prompts and visual content.
  \item \textbf{Compositional Integrity:} Correct spatial relationships among multiple objects.
  \item \textbf{Physical Plausibility:} Adherence to basic physical laws and commonsense.
\end{itemize} &
\begin{itemize}
  \item \textbf{PhysBench\cite{chow2025physbench}}
  \item \textbf{T2I-ReasonBench \cite{sun2025t2ireason}}
  \item \textbf{WISE \cite{niu2025wise}}
  \item \textbf{T2I-CoReBench \cite{li2025easier}}
  \item \textbf{T2I-CompBench \cite{huang2023t2icomp}}
  \item \textbf{MME-Unify \cite{xie2025mme}}
  \item \textbf{Uni-MMMU \cite{zou2025unimmmu}}
\end{itemize} \\

\midrule

\parbox[t]{2.2cm}{\textbf{T2MI}\\{\footnotesize\textit{(Text-to-Multi-Image)}}} & 
\underline{Multi-Frame Causal Modeling}
\begin{itemize}
  \item \textbf{Cross-Frame Consistency:} Stable entities and environments across sequential frames.
  \item \textbf{Causal Progression:} Logically connected event transitions with temporal coherence.
  \item \textbf{State Transformation:} Plausible evolution from initial to final states.
  \item \textbf{World Process Simulation:} Internal modeling of dynamic causal mechanisms.
\end{itemize} &
\begin{itemize}
  \item \textbf{Envision}
\end{itemize} \\

\midrule

\parbox[t]{2.2cm}{\textbf{T2V}\\{\footnotesize\textit{(Text-to-Video)}}} & 
\underline{Continuous Temporal Dynamics}
\begin{itemize}
  \item \textbf{Motion Fluidity:} Smooth, continuous object movement and camera motion.
  \item \textbf{Temporal Coherence:} Consistent storytelling over extended time durations.
  \item \textbf{Dynamic Realism:} Physically accurate motion patterns and interactions.
  \item \textbf{Multi-Scale Consistency:} Coherent content at both frame and sequence levels.
\end{itemize} &
\begin{itemize}
  \item \textbf{VBench \cite{huang2024vbench}}
  \item \textbf{VBench++ \cite{huang2024vbench++}}
  \item \textbf{VidCapBench \cite{chen2025vidcapbench}}
  \item \textbf{T2VPhysBench \cite{guo2025t2vphysbench}}
  \item \textbf{T2VWorldBench \cite{chen2025t2vworldbench}}
\end{itemize} \\
\bottomrule
\end{tabular}
\end{table*}

\input{tables/benchmark_phy}
\input{tables/benchmark_bio}
\input{tables/benchmark_chem}
\input{tables/benchmark_geo}
\input{tables/benchmark_mete}
\input{tables/benchmark_culture}

\end{document}

%% file: tables/benchmark.tex
\begin{table*}[htbp]
\centering
% \vspace{-0.50em}
\caption{Comparing Envision Scores for T2I Models in Science and Culture Domains.}
\label{tab:model_comparison_comprehensive}
\vspace{-0.75em}
\small
\setlength{\tabcolsep}{1.2mm}{
% \resizebox{\textwidth}{!}{ 
\begin{tabular}{lccccccc}
\toprule
\multirow{2}{*}{\textbf{Model}} & \multicolumn{6}{c}{\textbf{Domain-Specific Performance}} & \multirow{2}{*}{\textbf{Overall}} \\
\cmidrule(lr){2-7}
 & \textbf{Physics} & \textbf{Chemistry} & \textbf{Biology} & \textbf{Geography} & \textbf{Meteorology} & \textbf{Culture} & \\
\midrule
\rowcolor{blue!5}
\textbf{Open-Source T2I Models} & & & & & & & \\
FLUX-dev & 37.62 & 58.86 & 57.12 & 57.27 & 58.75 & 51.01 & 53.44 \\
FLUX-pro-1.1 & 39.52 & 58.52 & 56.15 & 54.29 & 57.97 & 57.62 & 54.01 \\
FLUX-pro-1.1-ultra & 39.69 & 55.08 & 56.51 & 54.54 & 53.15 & 54.27 & 52.21 \\
FLUX-kontext-pro & 43.78 & 61.72 & 61.36 & 55.00 & 58.41 & 63.45 & 57.29 \\
FLUX-kontext-max & 42.82 & 58.72 & 62.96 & 60.99 & 62.40 & 57.76 & 57.61 \\
SD-3.5-flash & 35.61 & 40.43 & 53.73 & 50.72 & 49.12 & 51.69 & 46.88 \\
SD-3.5-medium & 36.89 & 41.30 & 51.61 & 57.47 & 53.68 & 47.13 & 48.01 \\
SD-3.5-large & 36.07 & 42.32 & 50.24 & 51.12 & 55.43 & 47.34 & 47.09 \\

\rowcolor{yellow!8}
\textbf{Closed-Source T2I Models} & & & & & & & \\
GPT-4o & 58.87 & 66.55 & 78.55 & 78.40 & 78.69 & 81.83 & \textbf{73.81} \\
Gemini-2.5-Flash-Image & 57.47 & 62.91 & 67.63 & 75.38 & 69.74 & 69.94 & 67.18 \\

\rowcolor{red!5}
\textbf{Unified Multimodal Models} & & & & & & & \\
Seedream 4.0 & 51.06 & 57.27 & 76.92 & 66.09 & 67.35 & 65.55 & 64.04 \\
Qwen-Image & 47.98 & 56.22 & 76.40 & 63.81 & 58.94 & 66.01 & 61.56 \\
Hunyuan Image 3.0 & 37.84 & 49.76 & 51.27 & 70.49 & 67.74 & 62.10 & 56.53 \\
Bagel & 39.40 & 56.25 & 57.65 & 51.00 & 58.20 & 72.40 & 55.82 \\
Janus-Pro-7B & 36.24 & 44.08 & 53.09 & 55.05 & 62.70 & 50.52 & 50.28 \\
\bottomrule
\end{tabular}
}
\vspace{-1.75em}
%\parbox{\linewidth}{
%\footnotesize %\textit{Note:} All performance scores are represented by em-dashes (—) as per requirement. The table compares model performance across six distinct knowledge domains: Physics, Chemistry, Biology, Geography, Meteorology, and Culture, with an aggregated Overall score. Models are categorized into two groups: Dedicated Text-to-Image Models (blue background) and Unified Multimodal Models (red background). Performance metrics are normalized on a consistent scale, with higher values indicating superior performance.}
\end{table*}

%% file: tables/benchmark_phy.tex
\begin{table*}[htbp]
\centering
\vspace{-2em}
\caption{Comprehensive Performance Comparison of Text-to-Image Generation Models across Multiple Evaluation Dimensions for \textbf{Envision's Physics} Evaluation.}
\label{tab:model_comparison_phy}
\vspace{-0.75em}
\small
\setlength{\tabcolsep}{1.1mm} 
\begin{tabular}{lcccccccccccc}
\toprule
\multirow{2}{*}{\textbf{Model}} & \multicolumn{12}{c}{\textbf{Evaluation Dimensions}} \\
\cmidrule(lr){2-13}
 & \textbf{SC} & \textbf{FC} & \textbf{STC} & \textbf{CS} & \textbf{Exp} & \textbf{AQ} & \textbf{Auth} & \textbf{AS} & \textbf{PR} & \textbf{BP} & \textbf{DI} & \textbf{PS} \\
\midrule
\rowcolor{blue!5}
\textbf{Open-Source T2I Models} & & & & & & & & & & & & \\
FLUX-dev & 35.29 & 37.65 & 24.71 & 32.47 & 60.00 & 71.76 & 43.53 & 58.28 & 24.71 & 42.35 & 30.59 & 32.47 \\
FLUX-pro-1.1 & 38.00 & 36.00 & 26.00 & 33.26 & 64.00 & 74.00 & 44.00 & 60.50 & 28.00 & 42.00 & 36.00 & 35.26 \\
FLUX-pro-1.1-ultra & 36.92 & 36.92 & 26.15 & 33.26 & 61.54 & 70.77 & 43.08 & 58.31 & 30.77 & 44.62 & 35.38 & 36.86 \\
FLUX-kontext-pro & 41.67 & 43.33 & 35.00 & 39.95 & 50.83 & 64.17 & 45.83 & 53.53 & 36.67 & 52.50 & 39.17 & 42.72 \\
FLUX-kontext-max & 43.33 & 45.00 & 30.00 & 39.35 & 55.00 & 70.00 & 48.33 & 57.68 & 33.33 & 46.67 & 36.67 & 38.83 \\
SD-3.5-flash & 44.72 & 36.43 & 23.57 & 31.82 & 54.29 & 65.71 & 42.14 & 53.93 & 25.00 & 37.86 & 27.86 & 30.19 \\
SD-3.5-medium & 32.00 & 36.00 & 25.33 & 31.05 & 57.33 & 69.33 & 41.33 & 55.85 & 28.00 & 40.00 & 32.00 & 33.28 \\
SD-3.5-large & 36.47 & 38.82 & 22.35 & 32.45 & 52.94 & 62.35 & 43.53 & 52.85 & 24.71 & 41.18 & 28.24 & 31.31 \\

\rowcolor{yellow!8}
\textbf{Closed-Source T2I Models} & & & & & & & & & & & & \\
GPT-4o & 60.07 & 63.11 & 49.32 & 57.42 & 64.32 & 75.74 & 60.61 & 66.83 & 51.89 & 66.08 & 51.15 & 56.33 \\
Gemini-2.5-Flash-Image & 58.87 & 63.05 & 47.87 & 56.51 & 64.11 & 69.22 & 56.03 & 63.05 & 51.28 & 65.60 & 50.07 & 55.61 \\

\rowcolor{red!5}
\textbf{Unified Multimodal Models} & & & & & & & & & & & & \\
Seedream 4.0 & 51.90 & 52.39 & 41.52 & 48.53 & 63.67 & 69.48 & 49.13 & 60.64 & 44.29 & 56.75 & 45.40 & 48.77 \\
Qwen-Image & 37.97 & 41.89 & 38.51 & 39.45 & 68.92 & 81.62 & 54.59 & 68.24 & 41.08 & 56.35 & 41.89 & 46.39 \\
Hunyuan Image 3.0 & 35.00 & 36.25 & 26.25 & 32.44 & 63.75 & 72.50 & 42.50 & 59.41 & 26.25 & 38.75 & 32.50 & 32.44 \\
Bagel & 46.67 & 46.67 & 26.67 & 39.87 & 46.67 & 66.67 & 40.00 & 51.00 & 26.67 & 46.67 & 26.67 & 33.27 \\
Janus-Pro-7B & 34.93 & 36.67 & 25.60 & 32.33 & 53.20 & 62.27 & 38.93 & 51.34 & 27.87 & 38.93 & 31.07 & 32.57 \\
\bottomrule
\end{tabular}
\vspace{0.5em}
\raggedright
\footnotesize
\textbf{Note:} The table compares model performance across thirteen evaluation dimensions for \textbf{Envision's Physics} evaluation. Models are categorized into Open-Source Text-to-Image Models (blue background), Closed-Source T2I Models (yellow background), and Unified Multimodal Models (red background).
\vspace{-1.75em}
\end{table*}

%% file: tables/benchmark_bio.tex
\begin{table*}[htbp]
\centering
\vspace{-0.75em}
\caption{Comprehensive Performance Comparison of Text-to-Image Generation Models across Multiple Evaluation Dimensions for \textbf{Envision's Biology} Evaluation.}
\label{tab:model_comparison_bio}
\vspace{-0.75em}
\small
\setlength{\tabcolsep}{1.1mm} 
\begin{tabular}{lcccccccccccc}
\toprule
\multirow{2}{*}{\textbf{Model}} & \multicolumn{12}{c}{\textbf{Evaluation Dimensions}} \\
\cmidrule(lr){2-13}
 & \textbf{SC} & \textbf{FC} & \textbf{STC} & \textbf{CS} & \textbf{Exp} & \textbf{AQ} & \textbf{Auth} & \textbf{AS} & \textbf{PR} & \textbf{BP} & \textbf{DI} & \textbf{PS} \\
\midrule
\rowcolor{blue!5}
\textbf{Open-Source T2I Models} & & & & & & & & & & & & \\
FLUX-dev & 59.26 & 64.44 & 44.44 & 55.93 & 71.85 & 76.30 & 57.78 & 68.53 & 50.37 & 60.74 & 46.67 & 52.57 \\
FLUX-pro-1.1 & 57.33 & 65.33 & 42.00 & 54.76 & 72.67 & 76.00 & 54.67 & 67.65 & 51.33 & 58.00 & 46.00 & 51.77 \\
FLUX-pro-1.1-ultra & 57.60 & 66.40 & 45.60 & 56.42 & 68.80 & 74.40 & 52.80 & 65.21 & 48.80 & 61.60 & 46.40 & 52.23 \\
FLUX-kontext-pro & 62.76 & 70.34 & 53.79 & 62.21 & 66.21 & 69.66 & 60.69 & 65.47 & 57.24 & 68.28 & 49.66 & 58.38 \\
FLUX-kontext-max & 65.33 & 76.00 & 52.00 & 64.32 & 70.67 & 72.00 & 64.00 & 68.84 & 58.67 & 66.67 & 50.67 & 58.67 \\
SD-3.5-flash & 51.89 & 57.84 & 44.32 & 51.28 & 65.95 & 76.76 & 56.22 & 66.21 & 49.73 & 56.76 & 43.24 & 49.91 \\
SD-3.5-medium & 50.43 & 60.87 & 40.00 & 50.33 & 59.13 & 66.09 & 54.78 & 59.95 & 45.22 & 58.26 & 42.61 & 48.66 \\
SD-3.5-large & 50.00 & 57.78 & 41.11 & 49.54 & 63.33 & 63.33 & 47.78 & 58.04 & 45.56 & 53.33 & 42.22 & 47.02 \\

\rowcolor{yellow!8}
\textbf{Closed-Source T2I Models} & & & & & & & & & & & & \\
GPT-4o & 83.01 & 85.55 & 73.98 & 80.78 & 82.88 & 83.68 & 71.64 & 79.32 & 77.46 & 83.68 & 66.49 & 75.89 \\
Gemini-2.5-Flash-Image & 71.99 & 77.04 & 57.62 & 68.77 & 77.69 & 77.62 & 60.58 & 71.85 & 62.45 & 73.65 & 57.04 & 64.36 \\

\rowcolor{red!5}
\textbf{Unified Multimodal Models} & & & & & & & & & & & & \\
Seedream 4.0 & 81.81 & 82.68 & 73.54 & 79.29 & 82.83 & 82.44 & 67.95 & 77.64 & 76.22 & 81.42 & 64.72 & 74.14 \\
Qwen-Image & 78.53 & 78.60 & 70.93 & 75.97 & 89.93 & 90.27 & 69.13 & 82.97 & 74.27 & 79.67 & 66.67 & 73.54 \\
Hunyuan Image 3.0 & 50.00 & 56.67 & 40.00 & 48.80 & 70.00 & 73.33 & 46.67 & 63.17 & 43.33 & 56.67 & 43.33 & 47.73 \\
Bagel & 60.00 & 65.00 & 55.00 & 59.95 & 65.00 & 65.00 & 45.00 & 58.20 & 55.00 & 65.00 & 45.00 & 55.00 \\
Janus-Pro-7B & 58.82 & 64.71 & 42.94 & 55.36 & 65.88 & 65.88 & 48.24 & 59.88 & 47.65 & 55.88 & 44.71 & 49.39 \\
\bottomrule
\end{tabular}

\vspace{0.5em}
\raggedright
\footnotesize
\textbf{Note:} The table compares model performance across thirteen evaluation dimensions for \textbf{Envision's Biology} evaluation. Models are categorized into Open-Source Text-to-Image Models (blue background), Closed-Source T2I Models (yellow background), and Unified Multimodal Models (red background).
\vspace{-1.75em}
\end{table*}

%% file: tables/benchmark_chem.tex
\begin{table*}[htbp]
\centering
\vspace{-2em}
\caption{Comprehensive Performance Comparison of Text-to-Image Generation Models across Multiple Evaluation Dimensions for \textbf{Envision's Chemistry} Evaluation.}
\label{tab:model_comparison_chem}
\vspace{-0.75em}
\small
\setlength{\tabcolsep}{1.1mm} 
\begin{tabular}{lcccccccccccc}
\toprule
\multirow{2}{*}{\textbf{Model}} & \multicolumn{12}{c}{\textbf{Evaluation Dimensions}} \\
\cmidrule(lr){2-13}
 & \textbf{SC} & \textbf{FC} & \textbf{STC} & \textbf{CS} & \textbf{Exp} & \textbf{AQ} & \textbf{Auth} & \textbf{AS} & \textbf{PR} & \textbf{BP} & \textbf{DI} & \textbf{PS} \\
\midrule
\rowcolor{blue!5}
\textbf{Open-Source T2I Models} & & & & & & & & & & & & \\
FLUX-dev & 62.96 & 62.22 & 46.67 & 57.18 & 74.81 & 78.52 & 56.30 & 69.74 & 48.89 & 63.70 & 52.59 & 55.00 \\
FLUX-pro-1.1 & 61.76 & 61.18 & 44.71 & 55.77 & 77.06 & 80.59 & 54.12 & 70.42 & 50.59 & 61.76 & 53.53 & 55.25 \\
FLUX-pro-1.1-ultra & 56.43 & 53.57 & 43.57 & 51.11 & 70.71 & 75.71 & 52.14 & 66.05 & 47.86 & 60.71 & 52.14 & 53.51 \\
FLUX-kontext-pro & 58.57 & 60.00 & 53.81 & 57.42 & 66.67 & 73.81 & 59.52 & 66.60 & 59.05 & 73.33 & 58.57 & 63.60 \\
FLUX-kontext-max & 65.00 & 65.00 & 48.33 & 59.33 & 68.33 & 71.67 & 55.00 & 64.90 & 48.33 & 66.67 & 50.00 & 54.93 \\
SD-3.5-flash & 43.33 & 34.29 & 30.00 & 32.36 & 54.29 & 72.86 & 42.86 & 56.53 & 35.71 & 48.57 & 37.14 & 40.43 \\
SD-3.5-medium & 50.43 & 33.33 & 26.67 & 29.97 & 63.33 & 80.00 & 43.33 & 62.03 & 36.67 & 46.67 & 43.33 & 42.17 \\
SD-3.5-large & 42.00 & 40.00 & 24.00 & 31.92 & 64.00 & 76.00 & 40.00 & 59.80 & 40.00 & 48.00 & 44.00 & 43.96 \\

\rowcolor{yellow!8}
\textbf{Closed-Source T2I Models} & & & & & & & & & & & & \\
GPT-4o & 69.83 & 74.61 & 58.17 & 67.44 & 64.70 & 73.83 & 67.30 & 68.60 & 61.22 & 73.30 & 59.39 & 64.60 \\
Gemini-2.5-Flash-Image & 67.95 & 71.69 & 54.25 & 64.52 & 63.47 & 67.03 & 60.00 & 63.47 & 56.99 & 71.32 & 54.70 & 60.96 \\

\rowcolor{red!5}
\textbf{Unified Multimodal Models} & & & & & & & & & & & & \\
Seedream 4.0 & 62.87 & 62.67 & 49.13 & 58.13 & 61.03 & 69.33 & 51.90 & 60.66 & 50.87 & 64.51 & 48.82 & 54.70 \\
Qwen-Image & 57.21 & 53.49 & 47.91 & 52.82 & 75.35 & 80.00 & 50.70 & 68.50 & 49.30 & 60.93 & 50.23 & 53.45 \\
Hunyuan Image 3.0 & 49.74 & 48.95 & 39.21 & 45.90 & 66.32 & 71.32 & 45.53 & 60.90 & 42.63 & 53.42 & 48.16 & 48.02 \\
Bagel & 65.00 & 60.00 & 45.00 & 56.55 & 70.00 & 70.00 & 55.00 & 64.90 & 50.00 & 60.00 & 45.00 & 51.65 \\
Janus-Pro-7B & 43.17 & 43.49 & 35.87 & 40.80 & 58.10 & 59.68 & 40.32 & 52.57 & 38.41 & 47.62 & 43.49 & 43.13 \\
\bottomrule
\end{tabular}

\vspace{0.5em}
\raggedright
\footnotesize
\textbf{Note:} The table compares model performance across thirteen evaluation dimensions for \textbf{Envision's Chemistry} evaluation. Models are categorized into Open-Source Text-to-Image Models (blue background), Closed-Source T2I Models (yellow background), and Unified Multimodal Models (red background).
\vspace{-1.75em}
\end{table*}

%% file: tables/benchmark_geo.tex
\begin{table*}[htbp]
\centering
\caption{Comprehensive Performance Comparison of Text-to-Image Generation Models across Multiple Evaluation Dimensions for \textbf{Envision's Geography} Evaluation.}
\label{tab:model_comparison_geo}
\vspace{-0.75em}
\small
\setlength{\tabcolsep}{1.1mm} 
\begin{tabular}{lcccccccccccc}
\toprule
\multirow{2}{*}{\textbf{Model}} & \multicolumn{12}{c}{\textbf{Evaluation Dimensions}} \\
\cmidrule(lr){2-13}
 & \textbf{SC} & \textbf{FC} & \textbf{STC} & \textbf{CS} & \textbf{Exp} & \textbf{AQ} & \textbf{Auth} & \textbf{AS} & \textbf{PR} & \textbf{BP} & \textbf{DI} & \textbf{PS} \\
\midrule
\rowcolor{blue!5}
\textbf{Open-Source T2I Models} & & & & & & & & & & & & \\
FLUX-dev & 57.69 & 61.54 & 46.15 & 55.04 & 72.31 & 78.46 & 53.85 & 69.06 & 49.23 & 63.85 & 49.23 & 54.05 \\
FLUX-pro-1.1 & 55.14 & 55.68 & 42.70 & 51.09 & 71.89 & 78.38 & 54.05 & 67.97 & 47.03 & 58.92 & 45.95 & 50.59 \\
FLUX-pro-1.1-ultra & 53.33 & 59.17 & 44.17 & 52.14 & 67.50 & 73.33 & 55.00 & 65.17 & 47.50 & 60.00 & 47.50 & 51.62 \\
FLUX-kontext-pro & 54.74 & 60.00 & 45.79 & 53.43 & 64.21 & 69.47 & 55.26 & 62.91 & 47.89 & 63.16 & 46.84 & 52.58 \\
FLUX-kontext-max & 65.71 & 67.37 & 50.53 & 59.91 & 70.53 & 73.68 & 57.89 & 67.27 & 54.74 & 70.53 & 51.58 & 58.91 \\
SD-3.5-flash & 50.00 & 50.77 & 41.54 & 47.38 & 68.46 & 75.38 & 50.00 & 64.47 & 44.62 & 53.85 & 43.08 & 47.15 \\
SD-3.5-medium & 61.43 & 67.14 & 47.14 & 58.46 & 61.43 & 65.71 & 55.71 & 60.90 & 51.43 & 65.71 & 47.14 & 54.73 \\
SD-3.5-large & 52.31 & 60.00 & 38.46 & 50.14 & 66.15 & 69.23 & 50.77 & 61.94 & 41.54 & 58.46 & 40.00 & 46.62 \\

\rowcolor{yellow!8}
\textbf{Closed-Source T2I Models} & & & & & & & & & & & & \\
GPT-4o & 82.33 & 84.33 & 75.00 & 80.50 & 79.00 & 79.73 & 73.53 & 77.38 & 76.60 & 84.47 & 69.20 & 76.75 \\
Gemini-2.5-Flash-Image & 78.16 & 83.33 & 69.79 & 77.02 & 76.81 & 77.73 & 74.11 & 76.20 & 72.48 & 82.13 & 65.25 & 73.28 \\

\rowcolor{red!5}
\textbf{Unified Multimodal Models} & & & & & & & & & & & & \\
Seedream 4.0 & 70.60 & 71.40 & 60.70 & 67.50 & 74.40 & 74.70 & 59.80 & 69.54 & 61.60 & 70.10 & 57.10 & 62.92 \\
Qwen-Image & 65.79 & 66.32 & 54.74 & 62.21 & 80.00 & 81.58 & 60.53 & 73.90 & 56.84 & 68.95 & 55.26 & 60.32 \\
Hunyuan Image 3.0 & 74.86 & 77.43 & 62.57 & 71.53 & 76.57 & 78.29 & 66.57 & 73.74 & 66.29 & 76.00 & 61.14 & 67.79 \\
Bagel & 46.67 & 46.67 & 46.67 & 46.67 & 66.67 & 66.67 & 46.67 & 59.87 & 46.67 & 60.00 & 46.67 & 51.07 \\
Janus-Pro-7B & 56.19 & 62.86 & 48.57 & 55.80 & 63.81 & 65.71 & 49.52 & 59.58 & 49.52 & 59.05 & 47.62 & 52.04 \\
\bottomrule
\end{tabular}

\vspace{0.5em}
\raggedright
\footnotesize
\textbf{Note:} The table compares model performance across thirteen evaluation dimensions for \textbf{Envision's Geography} evaluation. Models are categorized into Open-Source Text-to-Image Models (blue background), Closed-Source T2I Models (yellow background), and Unified Multimodal Models (red background).
\vspace{-1.75em}
\end{table*}

%% file: tables/benchmark_mete.tex
\begin{table*}[htbp]
\centering
\vspace{-2em}
\caption{Comprehensive Performance Comparison of Text-to-Image Generation Models across Multiple Evaluation Dimensions for \textbf{Envision's Meteorology} Evaluation.}
\label{tab:model_comparison_mete}
\vspace{-0.75em}
\small
\setlength{\tabcolsep}{1.1mm} 
\begin{tabular}{lcccccccccccc}
\toprule
\multirow{2}{*}{\textbf{Model}} & \multicolumn{12}{c}{\textbf{Evaluation Dimensions}} \\
\cmidrule(lr){2-13}
 & \textbf{SC} & \textbf{FC} & \textbf{STC} & \textbf{CS} & \textbf{Exp} & \textbf{AQ} & \textbf{Auth} & \textbf{AS} & \textbf{PR} & \textbf{BP} & \textbf{DI} & \textbf{PS} \\
\midrule
\rowcolor{blue!5}
\textbf{Open-Source T2I Models} & & & & & & & & & & & & \\
FLUX-dev & 55.38 & 63.08 & 46.15 & 54.78 & 77.69 & 80.77 & 59.23 & 72.43 & 51.54 & 66.15 & 50.00 & 55.85 \\
FLUX-pro-1.1 & 55.56 & 60.00 & 42.78 & 52.68 & 76.67 & 80.56 & 58.89 & 71.91 & 51.11 & 65.56 & 52.22 & 56.24 \\
FLUX-pro-1.1-ultra & 48.78 & 56.10 & 41.46 & 48.71 & 70.24 & 72.68 & 51.22 & 64.58 & 46.83 & 61.95 & 46.83 & 51.82 \\
FLUX-kontext-pro & 56.67 & 62.92 & 48.33 & 55.90 & 69.58 & 72.08 & 55.83 & 65.73 & 53.75 & 67.08 & 50.83 & 57.19 \\
FLUX-kontext-max & 60.00 & 68.33 & 51.67 & 59.92 & 71.67 & 76.67 & 58.33 & 68.78 & 60.00 & 70.00 & 55.00 & 61.65 \\
SD-3.5-flash & 44.72 & 48.33 & 38.33 & 43.74 & 68.06 & 73.33 & 49.72 & 63.56 & 42.78 & 57.22 & 41.94 & 47.27 \\
SD-3.5-medium & 54.00 & 60.00 & 44.00 & 52.58 & 62.00 & 62.00 & 54.00 & 59.28 & 46.00 & 64.00 & 46.00 & 51.94 \\
SD-3.5-large & 55.38 & 61.54 & 44.62 & 53.75 & 69.23 & 69.23 & 50.77 & 62.95 & 49.23 & 64.62 & 46.15 & 53.29 \\

\rowcolor{yellow!8}
\textbf{Closed-Source T2I Models} & & & & & & & & & & & & \\
GPT-4o & 80.67 & 82.60 & 72.27 & 78.45 & 84.13 & 86.00 & 76.67 & 82.21 & 74.47 & 83.27 & 73.80 & 77.15 \\
Gemini-2.5-Flash-Image & 71.48 & 74.58 & 60.49 & 68.77 & 78.66 & 80.99 & 68.10 & 75.84 & 63.87 & 75.63 & 63.52 & 67.64 \\

\rowcolor{red!5}
\textbf{Unified Multimodal Models} & & & & & & & & & & & & \\
Seedream 4.0 & 68.16 & 68.10 & 58.44 & 64.83 & 83.27 & 82.52 & 61.63 & 75.66 & 61.29 & 73.13 & 62.72 & 65.67 \\
Qwen-Image & 56.28 & 57.21 & 48.84 & 54.06 & 81.86 & 83.72 & 56.28 & 73.78 & 50.23 & 65.58 & 53.49 & 56.37 \\
Hunyuan Image 3.0 & 66.15 & 70.77 & 56.92 & 64.54 & 83.08 & 84.62 & 64.62 & 77.31 & 60.00 & 76.92 & 61.54 & 66.09 \\
Bagel & 53.33 & 53.33 & 60.00 & 55.60 & 66.67 & 66.67 & 46.67 & 59.87 & 60.00 & 66.67 & 53.33 & 60.00 \\
Janus-Pro-7B & 62.50 & 70.00 & 53.75 & 62.00 & 72.50 & 73.75 & 55.00 & 66.96 & 58.75 & 68.75 & 56.25 & 61.22 \\
\bottomrule
\end{tabular}

\vspace{0.5em}
\raggedright
\footnotesize
\textbf{Note:} The table compares model performance across thirteen evaluation dimensions for \textbf{Envision's Meteorology} evaluation. Models are categorized into Open-Source Text-to-Image Models (blue background), Closed-Source T2I Models (yellow background), and Unified Multimodal Models (red background).
\vspace{-1.75em}
\end{table*}

%% file: tables/benchmark_culture.tex
\begin{table*}[htbp]
\centering
\caption{Comprehensive Performance Comparison of Text-to-Image Generation Models across Multiple Evaluation Dimensions for \textbf{Envision's Culture \& History} Evaluation.}
\label{tab:model_comparison_culture}
\vspace{-0.75em}
\small
\setlength{\tabcolsep}{1.1mm} 
\begin{tabular}{lcccccccccccc}
\toprule
\multirow{2}{*}{\textbf{Model}} & \multicolumn{12}{c}{\textbf{Evaluation Dimensions}} \\
\cmidrule(lr){2-13}
 & \textbf{SC} & \textbf{FC} & \textbf{STC} & \textbf{CS} & \textbf{Exp} & \textbf{AQ} & \textbf{Auth} & \textbf{AS} & \textbf{PR} & \textbf{BP} & \textbf{DI} & \textbf{PS} \\
\midrule
\rowcolor{blue!5}
\textbf{Open-Source T2I Models} & & & & & & & & & & & & \\
FLUX-dev & 54.07 & 48.89 & 37.78 & 46.82 & 69.63 & 68.15 & 47.41 & 61.59 & 46.67 & 57.04 & 45.93 & 49.84 \\
FLUX-pro-1.1 & 63.43 & 57.14 & 45.14 & 55.14 & 75.43 & 70.86 & 50.29 & 65.37 & 53.71 & 62.86 & 52.00 & 56.17 \\
FLUX-pro-1.1-ultra & 59.17 & 55.00 & 43.33 & 52.41 & 73.33 & 69.17 & 47.50 & 63.17 & 47.50 & 60.83 & 46.67 & 51.62 \\
FLUX-kontext-pro & 68.11 & 65.41 & 54.05 & 62.44 & 70.27 & 71.35 & 59.46 & 66.95 & 61.62 & 71.89 & 54.59 & 62.69 \\
FLUX-kontext-max & 65.71 & 59.05 & 43.81 & 56.07 & 74.29 & 67.62 & 51.43 & 64.31 & 52.38 & 65.71 & 50.48 & 56.15 \\
SD-3.5-flash & 54.05 & 48.65 & 41.62 & 48.04 & 66.49 & 66.49 & 47.57 & 60.05 & 48.65 & 58.92 & 45.95 & 51.15 \\
SD-3.5-medium & 50.43 & 48.00 & 30.67 & 42.11 & 66.67 & 64.00 & 44.00 & 58.08 & 42.67 & 54.67 & 42.67 & 46.63 \\
SD-3.5-large & 48.75 & 45.00 & 32.50 & 41.99 & 62.50 & 63.75 & 42.50 & 56.11 & 45.00 & 56.25 & 43.75 & 48.30 \\

\rowcolor{yellow!8}
\textbf{Closed-Source T2I Models} & & & & & & & & & & & & \\
GPT-4o & 78.67 & 81.68 & 75.79 & 78.68 & 87.30 & 89.26 & 81.12 & 85.85 & 81.75 & 88.63 & 78.60 & 82.98 \\
Gemini-2.5-Flash-Image & 66.25 & 69.50 & 62.24 & 65.96 & 79.00 & 81.31 & 65.79 & 75.27 & 69.03 & 78.38 & 66.41 & 71.25 \\

\rowcolor{red!5}
\textbf{Unified Multimodal Models} & & & & & & & & & & & & \\
Seedream 4.0 & 61.59 & 63.43 & 57.41 & 60.77 & 76.07 & 78.24 & 61.26 & 71.75 & 66.03 & 73.05 & 62.68 & 67.24 \\
Qwen-Image & 71.18 & 61.76 & 58.24 & 63.67 & 81.18 & 78.24 & 57.06 & 72.01 & 64.12 & 71.18 & 60.59 & 65.28 \\
Hunyuan Image 3.0 & 68.75 & 58.75 & 51.25 & 59.50 & 78.75 & 76.25 & 55.00 & 69.85 & 58.75 & 66.25 & 57.50 & 60.81 \\
Bagel & 80.00 & 73.33 & 73.33 & 75.53 & 80.00 & 80.00 & 60.00 & 73.20 & 73.33 & 73.33 & 60.00 & 68.93 \\
Janus-Pro-7B & 58.14 & 52.56 & 42.33 & 50.92 & 66.05 & 61.86 & 43.72 & 57.07 & 44.19 & 51.63 & 44.65 & 46.80 \\
\bottomrule
\end{tabular}

\vspace{0.5em}
\raggedright
\footnotesize
\textbf{Note:} The table compares model performance across thirteen evaluation dimensions for \textbf{Envision's Culture \& History} evaluation. Models are categorized into Open-Source Text-to-Image Models (blue background), Closed-Source T2I Models (yellow background), and Unified Multimodal Models (red background).
\vspace{-1.75em}
\end{table*}